\documentclass{article}

\usepackage{microtype}
\usepackage{graphicx}
\usepackage{sidecap}
\usepackage{subfigure}
\usepackage{booktabs} 
\usepackage{hyperref}
\usepackage{url}
\usepackage{helvet}  
\usepackage{courier}  
\usepackage{graphicx} 
\usepackage{bbm}
\usepackage{listings}
\usepackage{amsmath}
\usepackage{amssymb}
\usepackage{mathtools}
\usepackage{amsthm}
\usepackage[capitalize,noabbrev]{cleveref}
\usepackage{algorithm}
\usepackage{algorithmic}
\usepackage{multirow}
\usepackage{makecell}
\usepackage{booktabs}
\usepackage{newfloat}
\usepackage{listings}
\usepackage{hyperref}

\usepackage[accepted]{icml2023}

\usepackage{natbib}  
\usepackage{caption} 
\theoremstyle{plain}

\theoremstyle{definition}

\theoremstyle{remark}

\DeclareMathOperator*{\argmax}{argmax}

\usepackage[textsize=tiny]{todonotes}

\icmltitlerunning{Correcting Flaws in Common Disentanglement Metrics}

\begin{document}

\twocolumn[
\icmltitle{Correcting Flaws in Common Disentanglement Metrics}



\icmlsetsymbol{equal}{*}

\begin{icmlauthorlist}
\icmlauthor{Louis Mahon}{x,y}
\icmlauthor{Lei Shah}{z,y}
\icmlauthor{Thomas Lukasiewicz}{a,y}
\end{icmlauthorlist}

\icmlaffiliation{x}{Department of Informatics, University of Edinburgh, , United Kingdom}
\icmlaffiliation{y}{Department of Computer Science, University of Oxford, United Kingdom}
\icmlaffiliation{z}{Artificial Intelligence Institute of Beihang University}
\icmlaffiliation{a}{School of Informatics, TU Wien, Austria}

\icmlcorrespondingauthor{Louis Mahon}{oneillml@tcd.ie}

\icmlkeywords{Machine Learning, ICML}

\vskip 0.3in
]



\printAffiliationsAndNotice{}  

\begin{abstract}
Recent years have seen growing interest in learning disentangled representations, in which distinct features, such as size or shape, are represented by distinct neurons. Quantifying the extent to which a given representation is disentangled is not straightforward; multiple metrics have been proposed. In this paper, we identify two failings of existing metrics, which mean they can assign a high score to a model which is still entangled, and we propose two new metrics, which redress these problems. We then consider the task of compositional generalization. Unlike prior works, we treat this as a classification problem, which allows us to use it to measure the disentanglement ability of the encoder, without depending on the decoder. We show that performance on this task is (a) generally quite poor, (b) correlated with most disentanglement metrics, and (c) most strongly correlated with our newly proposed metrics.
\end{abstract}

\section{Introduction}
Learning to produce an effective vector representation of input data has long been a central question for the field of deep learning. Early proponents argued that a significant advantage of neural networks was that they could form distributed representations, where each input is represented by multiple neurons, and each neuron is involved in the representation of multiple different inputs \citep{hinton1986learning}. Compared to using a separate neuron for each input, distributed representations are exponentially more compact \citep{bengio2009learning}. An extension of distributed representations is the idea of disentangled representations. These are a particular type of distributed representation in which each neuron represents a single human-interpretable feature of the input data, such as colour, size or shape. In the disentanglement literature, these features are often referred to as `generative factors' or `factors of variation'.
Intuitively, a disentangled vector representation is one in which each factor of variation is represented by a distinct subset of neurons. E.g., a certain subset of neurons represents shape and shape only, another distinct subset represents size and size only etc, and changing the size of the input but not the shape, will mean that the size neurons change their activation value, but the shape neurons remain unchanged. The difficulty, and the subject of this paper, comes when we try to translate this intuition into quantifiable metrics.
In the strongest case, each factor is represented by a single neuron so that, e.g., changing the colour of the object in the image would cause a single neuron to change its value while all other neurons remain unchanged. (We discuss further below the ambiguity as to whether this stronger condition is required.) Disentanglement (DE) was originally formulated by \cite{bengio2009learning} (see also \citep{bengio2013better,bengio2013deep}). More recently, beginning with \citep{higgins2016beta}, there have been many unsupervised DE methods proposed based on autoencoders. 

In this paper, we examine the commonly used metrics to assess disentanglement. Firstly, we show how they fail to pick up certain forms of entanglement, and that representations can score highly on such metrics while being entangled. Specifically, we expose two problems with existing metrics: that they incorrectly align ground-truth factors to neurons, as they do not require distinct variables to be assigned to distinct factors; and that they only consider the effect of single latent variables at a time, and so fail to detect entanglements spread over multiple neurons. To address these problems, we present two new DE metrics, based on the ability of a classifier to predict the generative factors from the encoded representation. If a representation is truly disentangled, then all the relevant information should be contained in a single neuron (or possibly a few neurons, see discussion in Section \ref{sec:related-work}), and so a classifier using only this/these neuron(s) should be just as accurate as one using all neurons, and one using all other neurons should be very inaccurate. Our first metric is the accuracy of the single-neuron classifier, this should be high. Our second metric is the accuracy of the classifier using all other neurons (normalized by that of a classifier on all neurons), this should be low.

As well as showing the theoretical flaws in existing metrics, we also establish the superiority of our proposed metrics empirically, using a downstream compositional generalization task of identifying novel combinations of familiar features. Humans could clearly recognize a purple giraffe, even if we have never seen one or even heard the phrase ``purple giraffe'' before, because we have disentangled the concepts of colour and shape, so could recognize each separately. The ability to form and understand novel combinations is a deep, important aspect of human cognition
and is a direct consequence of humans being able to disentangle the relevant features of the objects we encounter. 
This is the basis for our proposed task. For example, we test whether a network trained to identify blue squares, blue circles and yellow squares can, at test time, correctly identify yellow circles. If it had learned to disentangle colour from shape, then it could simply identify ``yellow'' and ``circle'' separately, each of which is familiar. We show that existing DE models generally perform poorly at this task, suggesting they are further from DE than previous analyses have implied. We also show that a high score on DE metrics is predictive of performance on this task, and that our proposed DE metrics are the most predictive in this respect. 
Our contributions are briefly summarized as follows.
\begin{itemize}
    \item We identify and describe two shortcomings of existing DE metrics: incorrect alignment of neurons to factors and failure to pick on distributed entanglements. 
    \item We propose two alternative metrics, single-neuron classification and neuron knockout, that do not suffer from the problems that existing metrics suffer from.
    \item We introduce the task of identifying novel combinations of familiar features to measure the compositional generalization of the encoding, and describe why it is suitable for evaluating DE models. 
    \item We show empirically that existing models generally perform badly at this task, that their performance correlates with most DE metrics, and that the strongest correlation is with our proposed metrics.
\end{itemize}
The rest of this paper is organized as follows. Section \ref{sec:related-work} discusses related work. Section \ref{sec:problems-with-existing} describes the shortcomings of existing DE metrics, Section \ref{sec:PM} proposes our new metrics. Section \ref{sec:CG} introduces the task of classifying novel combinations. Section \ref{sec:experimental-eval} presents the results of applying our metrics and proposed task to existing DE models, and Section \ref{sec:conclusion} summarizes our work.

\section{Related Work} \label{sec:related-work}
\paragraph{Disentanglement Models.} 
\citet{higgins2016beta} proposed $\beta$-VAE, as an adaption of the variational autoencoder (VAE) \citep{kingma2013auto} to produce disentangled representations. By taking the prior to have a diagonal covariance matrix,  and increasing the Kullbeck-Leibler divergence (KL) loss weight, $\beta$-VAE encourages the model representations to have diagonal covariance too, which the authors claim enforces DE. 
\citet{kumar2017variational} further encouraged a diagonal covariance matrix by minimizing the Euclidean distance of the model's covariance matrix from the identity matrix. \citet{burgess2018understanding} proposed to gradually increase the reconstruction capacity of the autoencoder by annealing the KL in the VAE loss. Note that these works implicitly equate uncorrelated variables (i.e., diagonal covariance matrix) with independent variables, which is incorrect. For example, $Y = X^2$ are totally uncorrelated but totally dependent. \citet{chen2018isolating} propose $\beta$-TCVAE, which seeks to minimize the total correlation of the latent variables, that is, the KL divergence of the joint distribution from the product of the marginals. Total correlation is approximated using Monte-Carlo on a sampling method inspired by importance sampling. \citet{kim2018disentangling} propose FactorVAE, which also minimizes total correlation, this time approximating using a discriminator network. 
\citet{locatello2019challenging} challenged earlier DE methods, proving that it is always possible for a model to learn an entangled representation that appears disentangled only on the available data and presenting experiments suggesting that disentangled representations do not necessarily perform better on downstream tasks such as classification. Later, \citet{locatello2020weakly} claimed that including a small amount of supervision was sufficient to learn disentangled representations. Some recent DE methods are still unsupervised, \citet{klindt2020towards} proposed a VAE-based model to learn disentangled representations from videos of natural scenes. 
Semi-supervised methods include WeakDE \citep{valenti2022leveraging}, which employs adversarial training to encourage latent distribution for each factor to be close to a prior distribution computed using a small number of labelled examples. Some are fully supervised, such as MTD \citep{sha2021multi}, which partitions the neurons into a subset for each factor, and defines several additional DE loss terms using the ground-truth labels. Some can operate either supervised or unsupervised, such as Parted-VAE \citep{hajimiri2021semi}, which encourages DE by minimizing the Bhattacharyya distance~(\citeyear{Bhattacharyya1946}) from a multivariate normal.

\paragraph{Disentanglement Metrics.}
\citet{higgins2016beta} made an early attempt to quantify disentanglement, and proposed fixing one of the generative factors and varying all others, then using a linear model to predict which factor was fixed from the variance in each of the latent neurons. They claim that high classification accuracy implies better DE. \citet{kim2018disentangling} also propose a linear classifier, whose input is the index of the neuron with the lowest empirical variance.
\citet{ridgeway2018learning} decompose DE into two different concepts, modularity and explicitness, each with its own metric. They measure modularity as the deviation from an ``ideally modular'' representation, where every latent neuron has nonzero mutual information with exactly one generative factor. Explicitness is measured similarly to the metric of \citet{kim2018disentangling}, except using the mean of one-vs-rest classification and AUC-ROC instead of accuracy. The SAP score \cite{kumar2017variational} calculates, for each factor, the $R^2$ coefficient with each latent dimension, and then takes the difference between the largest and second largest. A representation will score highly on SAP if, for each generative factor, one neuron is very informative and no other individual neuron is. A similar idea is employed by the mutual information gap (MIG) metric \citep{chen2018isolating}, except using mutual information instead of $R^2$. \citet{eastwood2018framework} decompose the task into disentanglement, completeness and informativeness (DCI). Then, they train a classifier (most commonly a linear model or a gradient-boosted tree) to predict each generative factor from the latent factors, and estimate DCI from a measure of importance of each feature to each factor from the classifier. 
IRS \citep{suter2019robustly} measures the maximum amount that a given neuron can be changed by changing a generative factor other than the one it corresponds to. Another recent metric is MED \cite{cao2022empirical}, which is defined similarly to DCI, except designed to work effectively high-dimensional latent spaces.

\paragraph{Definitions of Disentanglement.} \label{sec:strong-weak-DE}
There is some ambiguity in the literature as to whether DE requires that each factor is represented by a single neuron, or allows representation by multiple neurons. The original definition by \citet{bengio2009learning} (echoed by \citet{higgins2016beta},  \citet{kim2018disentangling} and others) just stipulates that each neuron represents a single factor, and seems to allow that each factor is represented by multiple neurons. However, a stronger concept of DE is normally implied, namely, an injective function from factors to the unique neurons that represent them. This is implicit in the common metrics,  which map each factor to a single neuron, 
and in the descriptions of DE in other works, e.g., ``learning one exclusive factor per dimension'' \citep{pineau2018infocatvae}. Attempts at formal definitions, e.g. equivariance in group theory \citep{higgins2018towards}, also allow the possibility of many-to-one mappings in theory, but present examples and discuss benefits of DE with respect to one-to-one mappings. The weaker notion of DE, which allows the representation of a factor by multiple neurons, loses some of the claimed benefits of DE: it would be hard to tell which neurons represent a given factor if we have to check all \emph{subsets} of neurons, of which there are exponentially many, and we may still be unable to interpret what a single neuron represents, if all we know is what sets of neurons represent. Here, we mostly consider the strong notion, for the following reasons: (1) strong DE is assumed by all the commonly used DE metrics, and it is unclear how to quantify weak DE, (2) strong DE is assumed by the common practice of latent traversals, (3) weak DE loses interpretability as compared with strong DE, and it is debatable what advantages weak DE carries over an entangled representation.
It is mostly towards the strong notion that our challenges are directed, but the MTD method we test on in Section \ref{sec:experimental-eval} is based on weak DE, which allows us to see whether it performs any better on the compositional generalization task. 

\section{Problems with Existing Disentanglement Metrics} \label{sec:problems-with-existing}
\subsection{Incorrect Alignment of Latent Variables} \label{subsec:incorrect-alignment}
\begin{figure}
    \centering
    \includegraphics[width=0.4\textwidth]{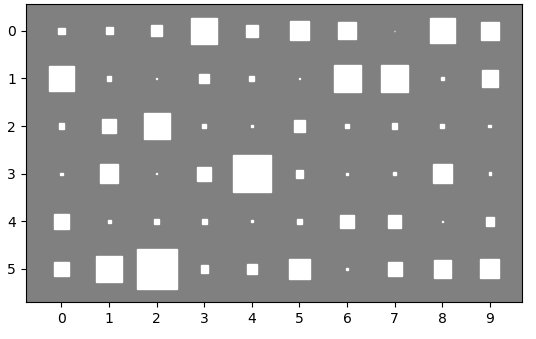}
    \vspace{-0.5em}
    \caption{Hinton diagram for $\beta$-TCVAE on 3dshapes. The size of the square at $(i,j)$ is proportional to the mutual information between factor $i$ and neuron $j$. Existing metrics incorrectly align both factor 2 and factor 5 to neuron 2, whereas we correctly align factor 2 to neuron 1 instead.}
    \label{fig:real-confused-alignment}
    \vspace{-8pt}
\end{figure}

The majority of existing metrics are based on aligning the set of factors $G$ with the set of neurons~$Z$; that is, for each factor, finding the neuron that it is represented by. Each neuron is only supposed to represent a single factor, however, all existing metrics simply relate each factor to the maximally informative variable, e.g., as measured by mutual information~\citep{chen2018isolating} or weight from a linear classifier~\citep{kumar2017variational} etc. This does not enforce the constraint of having distinct neurons for distinct features, it means that the same neuron could be selected as representing multiple different factors. For example, consider again the model trained on a dataset of blue squares, blue circles, yellow squares and yellow circles. Suppose such a model has two latent variables, $z_1$ and $z_2$, the first is related as per Table \ref{tab:double-encoding}, and the second is random noise, unrelated to the inputs. 
\begin{table}[]
    \centering
    \begin{tabular}{ccc}
\toprule[1.0pt]
         &  blue & yellow\\
         \midrule[0.5pt]
       Square  & 0 & \thead{0 or 1 with \\ equal probability} \\
       Circle  & \thead{0 or 1 with \\ equal probability} & 1 \\
       \bottomrule[1.0pt]
    \end{tabular}
    \caption{Toy example of partially encoding two factors.}
    \label{tab:double-encoding}
    \vspace{-1ex}

\end{table}
Then $z_1$ encodes both colour and shape, each to an accuracy of 75\%, whereas $z_2$ encodes both only to 50\% (random guess). Thus, $z_1$ will be chosen as the representative of both colour and shape. We observe that cases like this, where one neuron strongly encodes two or more factors, often occur in practice. Figure~\ref{fig:real-confused-alignment} gives a real-world example where the approach taken by existing metrics of aligning each factor to the neuron with the highest informativeness (e.g., by mutual information), incorrectly concludes that two different factors, size (factor 2) and colour (factor 5) should be aligned to the same neuron (neuron 2). Our method enforces distinct neurons, and so correctly aligns size to neuron 1 and colour to neuron 2. (Further similar examples in the appendix.) 

One may feel that Table \ref{tab:double-encoding} is not a problem, as that neuron only represents each of the two factors to 75\% accuracy, and we want a model that represents at close to 100\% accuracy. However, a metric should not only work properly for models that are close to perfect, it should also be able to distinguish between two models, both of which are far from perfect but one of which is better than the other. Incorrect alignment can interfere with distinguishing between such models. The model from Table \ref{tab:double-encoding}, would get a higher score than another in which $z_1$ is as above and $z_2$ represents shape to an accuracy of $70\%$. The appendix gives calculations of MIG, DCI and SAP in this case, which all decrease, and our metrics, which increase. Decreasing is incorrect, as the second model is clearly closer to the desired disentangled representation in which $z_1$ represents colour and colour only, and $z_2$ represents shape and shape only (WLOG on ordering of the neurons). 
The first model needs $z_2$ to learn shape and $z_1$ to unlearn shape, the second model just needs $z_1$ to unlearn shape. See also the appendix in \cite{cao2022empirical}, which discusses a similar failing.

Rather than aligning each factor independently, we should align all factors simultaneously and enforce that they are aligned to distinct neurons:
\begin{equation} \label{eq:correct-alignment}
    \argmax_{\{f:G \rightarrow Z| \text{ f is injective }\}} \sum_{g \in G} I(g;f(g)) \,,
\end{equation}
where $I$ denotes mutual information (this could be replaced with $R^2$ or any other measure of informativeness). A solution to \eqref{eq:correct-alignment} can be computed efficiently using the Kuhn-Munkres algorithm~\citep{munkres1957algorithms}. The result is a mapping from factors in $G$ to neurons in $Z$ where no two factors are mapped to the same neuron. This better fits the notion of DE than previous approaches.

\subsection{Distributed Entanglements} \label{subsec:distributed-entanglements}
The second problem with existing metrics is that they miss information that may be distributed over multiple neurons. 
In what follows, let $g_0, \dots, g_n$ be the ground-truth generative factors, and let $z = z_0, \dots, z_m$, $m \geq n$ be the corresponding neurons, ordered so that $z_i$ has been selected to correspond to $g_i$ for each $i$. Further, let $z_{\neq i}$ refer to the set of all neurons other than $z_i$. Now, consider XOR: $g_0 = z_1 \oplus z_2$. (We again use discrete variables for clarity, a continuous approximation could be $|z_1 - z_2|, z_1, z_2 \in [0,1]$.) Almost all existing metrics, both classifier-based and classifier-free, would conclude that $g_0$ is not represented by any other latent variable, and so may assign a high score.

For classifier-free metrics, the approach of existing methods is to compare individual neurons pairwise. The requirement is that, for each $i$, each individual neuron in $z_{\neq i}$ is uninformative about $g_i$. However, with XOR, this condition is met, but $z_{\neq i}$ collectively still (near) perfectly represent the $g_i$.  In this case, $I(g_0;z_1) = I(g_0;z_2) = 0$, but $I(g_0;z_1,z_2) \neq 0$, in fact $I(g_0;z_1,z_2)$ could be near maximal: $I(g_0;z_1,z_2) \approx H(g_0)$. 
Classifier-based metrics using linear classifiers would also fail to penalize $g_0 = z_1\oplus z_2$. There is insufficient information in this toy example to conclude what scores would be given by all classifier-based metrics using all (potentially non-linear) classifiers. The DCI score we investigate empirically in Section \ref{subsec:NCFF-experimental-eval} uses a gradient-boosted tree, which is the most common version, and still fails to predict the downstream task performance as well as our metrics do. There is also insufficient information to determine the IRS metric. However, this metric still incorrectly aligns factors (Section \ref{subsec:incorrect-alignment}), and we show the superiority of our metrics empirically in Section \ref{subsec:NCFF-experimental-eval}.

 Distributed entanglements are relevant not just to quantifying DE, but also to building DE models and losses. Early methods, such as $\beta$-VAE, made the mistake of equating uncorrelated with independent. Recent methods are more correct in focussing instead on information-theoretic measures of independence, but they still falsely equate pairwise independence with independence. We hope that redressing this mistake can lead to better DE models in future.

\section{Proposed Metrics}\label{sec:PM}

\paragraph{Single-neuron Classification (SNC)} 
Let $X = (x)_{1\leq i \leq N}$ denote the data, $c:X \rightarrow \{0, \dots, K-1\}$ specify the ground truth labels, and and $b:X \rightarrow \{0, \dots, K-1\}$ specify the bin index after alignment. Then
\[
SNC = \max(0,\frac{a-r}{1-r})\,,
\]
where 
\begin{gather}
a = \frac{1}{N}\sum_{i=1}^N \mathbbm{1}(b(x_i) = c(x_i)) \\
r = \frac{1}{N^2} \sum_{j=1}^K (\sum_{i=1}^N \mathbbm{1}(c(x_i) = j))^2 \,.
\end{gather}

This is essentially a quantification of the property that latent traversals aim to show qualitatively. In the terminology of \citet{ridgeway2018learning}, it is a measure of explicitness, except that they fit a linear classifier on all neurons, not just $z_i$, which therefore allows for non-axis-alignment. That is, if for each $g_i$, there is some line in representation space such that the representation vector encodes $g_i$ as the distance of the projection along that line, then this is regarded as disentangled. Using a single-neuron classifier, on the other hand, requires that line to be an axis. Section~\ref{subsec:NCFF-experimental-eval} also reports results for such a linear classifier.

\paragraph{Neuron Knockout (NK)} 
Our second proposed metric is inspired by the technique of gene knockout in genetics, which tests how relevant a given gene is to a given function, by removing the gene and measuring the loss in function. We test whether $z_{\neq i}$ contains any information about $g_i$, by training an MLP to predict $g_i$ from $z_{\neq i}$. If the representation is disentangled, then this accuracy should be low. By comparing to the accuracy of an MLP that uses all neurons, we can get an indication of how relevant $z_i$ is to representing $g_i$. Our second metric is $NK_i = Acc_{z} - Acc_{z_{\neq i}}$, where $A_{x}$ denotes the accuracy of an MLP trained on neurons $x$ to predict $g_i$. This is crucially different from most existing methods, which only test whether each neuron individually contains information about $g_i$, and which consequently miss distributed entanglements (Section \ref{subsec:distributed-entanglements}). Existing methods, e.g., fail to detect when $z_{\neq i}$ encode $g_i$ via XOR. Our method, however, because it uses an MLP, can detect non-linear encodings such as XOR. The results from Section \ref{sec:experimental-eval} show that such encodings occur in practice. 

Some existing works have used a similar idea \citep{sha2021multi,mathieu2016disentangling}. These works partition the set of neurons, a priori, into two subsets, $A$ and $B$, representing distinct factors $a$ and $b$, respectively. This is encouraged by using labelled examples. Performance is then measured by training an MLP to predict $b$ from $A$ and $a$ from $B$. This technique, however, is only applicable where the set of neurons has been partitioned during training by the use of labels, whereas ours is applicable to the case where we do not know which neurons represent which factors, because we include a method for identifying which neurons to knock out. Secondly, we measure the difference with an MLP trained on all neurons. This is important because it distinguishes between a disentangled representation, and one that simply contains no information about the input. For prior works, the latter would score highly, but for our NK metric, it would not. For such a representation, $Acc_{z} \approx Acc_{z_{\neq i}} \approx 0$.

\begin{table*}
\caption{Central tendency across five runs for our proposed metrics, SNC and NK, along with the accuracy predicting each factor using an MLP on all neurons (MLP) and a linear classifier on all neurons (linear). The best accuracy in each block is in bold.} \label{tab:SNC-SK-results}
\centering
\resizebox{0.9\textwidth}{!}{\begin{tabular}{llllllll}
\toprule
      &         &        $\beta$-VAE &        $\beta$-TCVAE &       FactorVAE &          PartedVAE &       PartedVAE-ss &        weakde \\
\midrule
dsprites & SNC &  24.8 (3.20) &   41.3 (3.00) &  15.1 (1.78) &   16.1 (4.54) &  19.0 (4.80) &    5.8 (0.75) \\
      & linear &  46.2 (2.54) &   49.5 (1.60) &  30.8 (2.43) &   36.0 (2.58) &  26.3 (10.03) &   28.6 (2.00) \\
      & MLP &  \textbf{89.4 (3.95)} &   \textbf{86.2 (1.10)} &  \textbf{77.1 (4.10)} &  \textbf{70.7 (10.33)} &  \textbf{48.5 (9.08)} &   \textbf{99.8 (3.50)} \\
      & NK &  39.9 (1.11) &   31.3 (4.10) &  32.9 (1.45) &   34.9 (4.23) &  14.5 (8.17) &    5.9 (1.17) \\
\midrule
3dshapes & SNC &  19.9 (4.70) &   20.1 (2.06) &  16.9 (3.36) &  43.8 (24.97) &   68.3 (9.55) &   15.4 (2.11) \\
      & linear &  79.0 (1.60) &   76.3 (3.47) &  65.2 (4.74) &   70.8 (6.73) &  65.4 (11.40) &  73.3 (18.17) \\
      & MLP &  \textbf{99.8 (3.50)} &   \textbf{99.9 (2.50)} &  \textbf{98.1 (7.24)} &   \textbf{96.2 (6.83)} &  \textbf{88.9 (14.53)} &   \textbf{99.8 (0.68)} \\
      & NK &   8.0 (0.20) &    8.1 (0.11) &  16.8 (2.94) &   60.2 (6.19) &   41.8 (8.07) &    2.5 (0.18) \\
\midrule
mpi3d & SNC &  32.6 (5.63) &   35.2 (4.43) &  27.5 (3.63) &   35.1 (3.29) &   23.9 (1.77) &   17.1 (0.39) \\
      & linear &  51.3 (2.73) &   52.1 (4.20) &  46.9 (2.52) &   49.4 (2.32) &   43.0 (5.11) &   46.2 (1.57) \\
      & MLP &  \textbf{91.3 (3.36)} &   \textbf{83.5 (9.00)} &  \textbf{88.6 (2.85)} &   \textbf{65.5 (1.51)} &   \textbf{60.6 (3.04)} &   \textbf{83.4 (0.89)} \\
      & NK &  19.8 (0.39) &  17.8 (10.82) &  23.2 (0.41) &    9.5 (1.64) &    3.3 (7.11) &    6.0 (0.14) \\
\bottomrule
\end{tabular}
}

\end{table*}

\section{Compositional Generalization by Recognizing Novel Combinations} \label{sec:CG}
Compositional generalization (CG) is the ability to combine familiar, learned concepts in novel ways.  Here, we cast CG as a classification task, and test whether the representations produced by a model can be used to correctly classify novel combinations of familiar features. 

If the representation learnt by a model is truly disentangled, then such novel combinations should not present any difficulty, because the model would be able to extract representation elements for each feature separately, and when considered separately, each is familiar to the model. However, if there is entanglement between the different features, then the novel combination is out of distribution. Formally, let $p(X,Y)$ be the empirical distribution from the data for the two features, $q(X,Y)$ the distribution learnt by the model, and $x$, $y$ the two feature values that do not appear together in the data (note that VAEs are generative models, so learn a joint distribution). Then $p(X=x)>0$, $p(Y=y) > 0$, but $p(X=x,Y=y) = 0$.
So, if $q$ has learnt a joint distribution, then $q(X=x,Y=y) \approx 0$, but if $q$ has been learned as a factored distribution, where $q(X) \approx p(X)$ and $q(Y) \approx p(Y)$, then $q(X=x,Y=y) = q(X)q(Y) > 0$. Our proposed task is as follows: 
\begin{enumerate}
    \item randomly sample values for two features, e.g., shape and size
    \item form a test set of points with those two values for those two features, e.g., all points with size=0 and shape=`square', and a train set of all other points
    \item  train the VAE (or supervised model) on the train set
    \item encode all data with the VAE encoder
    \item train and test an MLP on the train/test sets using the encodings as input and the generative factors as labels
\end{enumerate}

Previous works have noted the theoretical connection between CG and DE, and explored it in various ways. E.g., \citep{zhang2022equivariant} proposed a supervised classification model with regularization to enforce DE and assist classifying novel combinations. Other explorations of DE and CG are discussed in Section \ref{subsec:NCFF-experimental-eval}.


\section{Experimental Evaluation} \label{sec:experimental-eval}

\paragraph{Datasets.}
We test our metrics and task on three datasets. \textbf{Dsprites} contains 737,280 black-and-white images with features $(x,y)$-coordinates, size, orientation and shape. \textbf{3dshapes} contains 480,000 images with features object/ground/wall colour, size, camera azimuth, and shape. \textbf{MPI3D} contains 103,680 images of objects at the end of a robot arm with the features object colour, size and shape, camera height and azimuth, and altitude of the robot arm.

\paragraph{Implementation Details.}
We test several popular DE models, $\beta$-VAE, FactorVAE, $\beta$-TCVAE, PartedVAE and WeakDE. PartedVAE can be trained unsupervised or semisupervised. We test both settings. The details of these models are given in Section \ref{sec:related-work}. Parted-VAE and Weak-DE are trained using the original code at \url{https://github.com/sinahmr/parted-vae}, and \url{https://github.com/Andrea-V/Weak-Disentanglement}, respectively, for 100 epochs. Other models are trained using the library at \url{https://github.com/YannDubs/disentangling-vae}, all use default parameters. 
The MLPs and linear models trained to predict the factors are trained using Adam, learning rate $.001$, $\beta_1$=0.9, $\beta_2$=0.999, for 75 epochs. The MLP has one hidden layer of size $256$. MTD is trained using the author's code (obtained privately) with all default parameters, for 10 epochs.


\subsection{SNC and NK Results} \label{subsec:SNC-NK-experimental-eval}
Table \ref{tab:SNC-SK-results} shows the results of our two proposed metrics, SNC and NK on the three datasets described above. Each dataset includes a slightly different set of features, and displaying all features impairs readability, so we 
report the average across all features. Full results are given in the appendix.  

The SNC accuracy is substantially lower than that of the full MLP. The linear model performs slightly better, but is still significantly below the accuracy of the MLP. This suggests that each $g_i$ is represented more accurately by a distributed, non-linear entangled encoding across all neurons, rather than just by $z_i$. 
Although an MLP is a more powerful model, this should not help test set accuracy unless there is relevant information in the input that it can leverage, i.e., distributed entanglements. The fact that MLP accuracy is much higher than linear accuracy and that of SNC, suggests that such entanglements exist. Distributed entanglements are also suggested by the NK results. Here, there is often only a marginal drop in accuracy after removing the neuron that was supposed to contain all the relevant information, suggesting that much of the representation of $g_i$ is distributed over $z_{\neq i}$. As well as being evidence for entanglement, this highlights that fault tolerance is at odds with DE. In one sense, it is a good thing that after removing any one neuron, even that with the highest mutual information with the target feature, the representation still contains enough information to classify with reasonable accuracy. Fault-tolerance was one of the original arguments for the advantages of distributed representations \citep{fahlman1987connectionist,hinton1986learning}, and the brain has been shown to be highly fault-tolerant \citep{yu2016fault}. However, if, as disentangled models aim for, the information for each feature is contained in only one neuron, then the representation is very fragile, losing just one neuron means all the information is lost. This conceptual conflict with fault-tolerance suggests that DE may not always be desirable, and we should be clear about what benefit DE offers if we are to pursue it. The results from Table \ref{tab:SNC-SK-results} show that, to some extent, current supposedly disentangled models retain the benefit of fault-tolerance, at the cost of entanglement.



\subsection{Compositional Generalization Results} \label{subsec:NCFF-experimental-eval}
Table \ref{tab:CG-results} shows results for the task of classifying novel combinations of familiar features. As well as the models from Section \ref{subsec:SNC-NK-experimental-eval}, we also report results for MTD, a recent fully supervised method \citep{sha2021multi}. 

\begin{table*}[h]
\caption{Classification accuracy for novel combinations of familiar features, using an MLP and linear model, denoted `CG' and `CG linear' respectively, and classification accuracy when the test set is chosen randomly, denoted `normal test set'.} \label{tab:CG-results}
\centering
\resizebox{0.9\textwidth}{!}{
\begin{tabular}{llrrrrrrrrr}
\toprule
    &                & \multicolumn{3}{l}{dsprites} & \multicolumn{3}{l}{3dshapes} & \multicolumn{3}{l}{mpi3d} \\
    &                &    shape &   size &  both &    shape &   size &   both & shape &  size &  both \\
\midrule
$\beta$-VAE & CG &     0.00 &  61.55 &  0.00 &    33.00 &  67.70 &  14.75 & 89.00 &  3.66 &  2.34 \\
    & linear CG &     0.00 &   0.00 &  0.00 &    13.70 &  24.60 &   0.00 & 60.56 &  0.00 &  0.00 \\
    & normal test set &    82.35 &  66.24 & \textbf{89.44} &    96.30 &  96.90 &  \textbf{99.85} & 90.53 & 77.44 & \textbf{91.29} \\
\midrule
$\beta$TCVAE & CG &     0.00 &  49.56 &  0.00 &    29.40 &  77.50 &  17.73 & 87.87 &  0.00 &  0.00 \\
    & linear CG &     0.00 &   4.63 &  0.01 &     4.00 &  58.30 &   0.77 & 75.55 &  0.00 &  0.00 \\
    & normal test set &    82.32 &  66.22 & \textbf{86.21} &    96.50 &  96.80 &  \textbf{99.89} & 89.22 & 72.91 & \textbf{83.47} \\
\midrule
FactorVAE & CG &     0.00 &  37.28 &  0.86 &     0.86 &   6.36 &   6.36 & 89.36 &  0.47 &  0.00 \\
    & linear CG &     0.00 &   0.00 &  0.31 &    10.40 &  33.80 &   0.00 & 57.59 &  0.00 &  0.00 \\
    & normal test set &    80.85 &  65.08 & \textbf{77.05} &    89.50 &  95.70 &  \textbf{98.09} & 89.52 & 73.94 & \textbf{88.59} \\
\midrule
PartedVAE & CG &     0.00 &  29.61 &  0.00 &     0.00 &  19.11 &  18.52 & 81.28 & 91.67 &  0.00 \\
    & linear CG &     0.00 &   0.00 &  0.00 &     6.80 &  84.20 &   6.61 & 70.10 & 91.67 &  0.00 \\
    & normal test set &    58.33 &  63.83 & \textbf{70.65} &    93.50 &  95.90 &  \textbf{96.24} & 72.36 & 91.67 & \textbf{65.49} \\
\midrule
PartedVAE-ss & CG &     0.00 &  31.83 &  0.00 &     0.00 &  30.33 &  30.33 & 72.30 &  0.00 &  0.00 \\
    & linear CG &     0.00 &   0.00 &  0.00 &     3.70 &  33.40 &   0.00 & 60.80 &  0.00 &  0.00 \\
    & normal test set &    39.47 &  38.48 & \textbf{48.48} &    76.70 &  94.20 &  \textbf{88.93} & 71.20 & 25.70 & \textbf{60.61} \\
\midrule
WeakDE & CG &     0.00 &  41.79 &  0.00 &     0.00 &   7.83 &   7.83 & 87.10 &  0.00 &  0.00 \\
    & linear CG &     0.00 &   0.00 &  0.00 &     0.00 &  40.40 &   0.00 & 37.70 &  0.00 &  0.00 \\
    & normal test set &    74.66 &  65.58 & \textbf{79.01} &    96.50 &  96.90 &  \textbf{99.81} & 82.90 & 50.79 & \textbf{83.36} \\
\midrule
MTD & CG &     0.00 &   0.03 &  0.00 &     0.00 &   0.50 &   0.00 & 23.30 &  0.00 &  0.00 \\
    & normal test set &    99.66 & 100.00 & \textbf{99.56} &   100.00 & 100.00 & \textbf{100.00} & 99.80 & 95.01 & \textbf{95.36} \\
\bottomrule
\end{tabular}
}

\end{table*}

Because there is a very large number of combinations of feature values, it is not feasible to test all of them in enough detail to obtain reliable results. We restrict our attention to just shape and size combinations. We follow the procedure described in Section \ref{sec:CG}, using five feature combinations for dsprites, six for mpi3d and eight for 3dshapes (see appendix for details), and reporting the average. The ``normal test set'' setting uses the same method except divides the train and test sets randomly. We adjust for chance agreement as $\max(0,(a-r)/(1-r))$, where $a$ is the model accuracy, and $r$ is the chance agreement. Even restricting our attention to a single combination of feature types, our experimental results are already 
extensive involve training several hundred VAE models, and over 1000 classification heads (see appendix). An empirical study of multiple feature types would require 10x to 100x more, which would be a valuable future contribution, but is outside the scope of the present work. 

There is perhaps a danger that the MLP itself entangles two features that are not entangled in the representation itself. For example, if yellow circles are excluded then, when classifying shape, the MLP could learn that whenever the ``colour'' neuron indicates ``yellow'', it should place low probability mass on ``circle''. We feel this is unlikely to affect results significantly, as the relationship between the ``size'' neuron and the value of shape would be highly non-linear and present only for a small subset of data points. However, to ensure that this is not affecting performance, we also include another setting using a linear classifier instead of an MLP, so this non-linear relationship could not be learnt.

The accuracy for identifying novel combinations of familiar features is generally low, often at the level of random guessing (i.e., 0 after adjusting for chance agreement). This is even true for MTD, the supervised model. Under the `normal test set'' setting, every model is capable of classifying the unseen data accurately, which shows that it is the novelty of the combination that is degrading performance. The performance of the linear classifier is even worse, almost never above a random baseline, which suggests the poor performance is not due to the MLP learning a spurious correlation between these factors. This correlation is non-linear, so cannot be learnt by the linear classifier. If it were the cause for poor performance, we would therefore expect the linear model to perform better, which is not the case. 

There is a clear difference between the difficulty of the three datasets: results on dsprites and mpi3d are essentially always at the level of random guessing, whereas those on 3dshapes are more promising, reaching nearly 30\% (chance adjusted) for some models. Both variants of PartedVae show the best performance, with the semi-supervised variant being the best, though perhaps surprisingly, the other recent semi-supervised method, WeakDE, does not perform well. 


In machine learning, CG has mostly been studied in the context of language \citep{baroni2020linguistic}, and only rarely been connected to DE \citep{zheng2021disentangled}. In the field of DE, some prior works have claimed their model can meaningfully represent novel combinations: \citet{higgins2016beta} display figures of reconstructed chairs with a round bottom for certain latent traversals and \citet{esmaeili2019structured} present reconstructions for MNIST digits with certain combinations of digits and features (e.g., line thickness) excluded. Conflicting results were found by \citet{montero2020role}, who claimed to find that DE models almost entirely failed at representing novel combinations, and that performance was unrelated to the degree of DE itself. However, these prior works have only examined the reconstructions by the decoder and so do not provide sufficient evidence to make claims about the internal representation. Being able to reconstruct an image accurately does not establish anything about the internal representation, it could just be the result of learning the identity function. The key question is not what the decoder reconstructs from the encoding or from a latent traversal, it is what representation the encoder produces. Prior investigations into CG only examined the former.
Also, the experiment by \citet{montero2020role} attempted to quantify CG performance as the decoder's pixel loss. Not only does this depend on the quality of the decoder, not just the encoder, but pixel loss has long been observed to be a poor measure of the quality of a generated image \cite{oprea2020review,higgins2016beta}. Our proposed task of classifying novel combinations, on the other hand, more directly investigates what information is present in the encoder's representations of the data, and our experiments have more accurate measures of DE. Unlike \cite{higgins2016beta,esmaeili2019structured}, our experiments mostly show poor performance of existing DE models at CG. They also differ from \cite{montero2020role}, as they reveal a correlation between DE and CG, as shown in Section \ref{subsec:correlation}.

\vspace{-0.79ex}
\subsection{Our Metrics Predict Compositional Generalization} \label{subsec:correlation}
Disentangled representations are hoped to improve performance on some downstream tasks (or interpretability or efficiency). We have argued that identifying novel combinations of familiar features, qua compositional generalization, is such a task. It is costly to measure this directly, because it requires training multiple different models from scratch with different feature combinations excluded, but good DE metrics should be predictive of performance on this task. Here, we show that our metrics are more predictive of compositional generalization performance than are existing metrics.

Table \ref{tab:correls} shows the Pearson correlation coefficient of our metrics and existing metrics, computed from \url{https://github.com/google-research/disentanglement_lib}, with performance on the CG task from Section \ref{sec:CG}. When computing this correlation, we restrict our metrics to just size and shape, because they are the relevant features for this task. We also tried restricting existing metrics to these two features only, without much change in results; see appendix. There is no insight to be gained from the correlation on dsprites only or mpi3d only, since performance there is never more than slightly above random, so all correlations will be essentially zero. We show correlation on 3dshapes, and across all datasets. 
\begin{SCtable}[]
\begin{tabular}{lrr}
\toprule
{} &  all datasets &  3dshapes \\
\midrule
SNC &         \textbf{0.850} &     \textbf{0.849} \\
NK  &         \textit{0.716} &     0.710 \\
MIG &         0.571 &     0.800 \\
SAP &         0.311 &     0.726 \\
IRS &         0.106 &     0.668 \\
D   &         0.457 &     0.472 \\
C   &         0.414 &     \textit{0.819} \\
I   &         0.535 &     0.740 \\
DCI &         0.493 &     0.717 \\
MED &         0.566 &     0.785 \\
\bottomrule
\end{tabular}
\caption{Correlation (Pearson) of our metrics, and existing metrics, with accuracy on novel combinations. Best results in bold, second best italicized.} \label{tab:correls}
\vspace{-1.5em}
\end{SCtable}

The first observation is that all metrics are at least weakly correlated with compositional generalization performance, with some showing a moderate to strong correlation. This contradicts the claims of \citet{montero2020role}, who claimed to find no relationship between DE and CG. As argued in Section \ref{subsec:NCFF-experimental-eval}, our method for measuring CG performance is more indicative of the encoding quality. Additionally, we test six models, eight metrics and three datasets, whereas \citet{montero2020role} test only two models, one metric and two datasets. The consistent correlation between DE and CG evidenced by Table \ref{tab:correls} also validates our arguments, from Section \ref{sec:CG}, that DE should improve CG performance, and our choice to use CG to investigate DE quality.

Our SNC metric shows the highest correlation both on 3dshapes and on all datasets taken together. In both cases, the correlation is also statistically significant ($p<0.01$); calculation in the appendix. NK is also strongly correlated with compositional generalization, but less so than SNC. This is expected, because SNC measures the extent to which information about the generative factors is encoded in a disentangled way, whereas NK penalizes representations that also encode information in a redundant, entangled way. NK is perhaps more relevant than SNC to interpretability, where we want to know that a given factor is encoded by a single neuron \emph{only}, whereas SNC is more relevant for downstream performance. A similar comparison can be made between the components of the DCI metric, where the C and I components measure whether the information is present in a disentangled way, whereas the D component measures whether there is also information present in a disentangled way. These are analogous, modulo the superiority of our metrics, as discussed above, to NK and SNC, respectively, and, like NK, the D component is a less good predictor of downstream performance.

\vspace{-1.99ex}
\section{Conclusion} \label{sec:conclusion}
This paper has identified two common flaws in existing disentanglement metrics: incorrect alignment of generative factors with neurons and a failure to detect distributed entanglement. We introduced two new metrics, single-neuron classification and neuron knockout, which avoid these problems. Next, we propose the classification of novel combinations of familiar features as a real-world task with which disentanglement can be measured. We show that our metrics are strongly predictive of performance on this task, more strongly than are existing metrics.


\bibliography{bibliography}

\clearpage

\begin{appendix}
\section{Calculations from the Examples in Table \ref{tab:double-encoding}}
Here, we show the calculation of MIG, DCI, and SAP, as well as our metrics, from the example of Table \ref{tab:double-encoding} in Section \ref{sec:problems-with-existing}. We are interested in the change of these metrics when $z_1$The others all, incorrectly, decrease, while ours, correctly, increase.

\paragraph{MIG.}
Let $g_1$ denote shape and $g_2$ denote colour, and let $H$ denote entropy and $MI$ denote mutual information. We then calculate the MIG for the model in Table 1 as follows. First, note that
\begin{gather}
    MI(z_2,g_1) = MI(z_2,g_2) = 0 \,,
\end{gather}
because $z_2$ is just noise. Then, assuming balanced classes, we have $H(g_1) = H(g_2) = 1$. As $z_1$ encodes each to an accuracy of $75\%$, the conditional entropy is 
\[
-0.75 \log(0.75) - (0.25) \log{0.25} = 0.8113\,,
\]
and so
\[
MI(z_1,g_1) = MI(z_1,g_2) = 1 - 0.8113 = 0.1887\,.
\]
The MIG, which is identical for both features, is then 
\begin{gather*}
MI(z_1,g_1) - MI(z_2,g_1) = MI(z_1,g_2) - MI(z_2,g_2) =  \\ 0.1887 - 0 = 0.1887\,.
\end{gather*}

Now, we calculate MIG for the second described model. Here,
\[
MI(z_1,g_1) = MI(z_1,g_2) = 0.1887\,,
\]
as above, and also $MI(z_2,g_1)=0$ as above. Now, however,
\begin{gather*}
MI(z_2,g_2) = 1 - (-0.7 \log(0.7) - (0.3) \log{0.3}) =  \\ 1 - 0.8813 = 0.1187\,,
\end{gather*}
So, the MIG for $g_1$ is the same as for the first model, but the MIG for $g_2$ is
\[
MI(z_1,g_2) - MI(z_2,g_2) = 0.1887 - 0.1187 = 0.07\,.
\]

\paragraph{DCI.}
The $I$ component is unchanged at $75\%$. To calculate $D$ and $C$, we need some measure of the importance or strength of the connection of each neuron to each feature. DCI usually measures feature importance with a classifier, but of course we cannot train a classifier on a theoretical example. The only information is the accuracy to which each neuron encodes each feature, and this provides a reasonable way to quantify feature importance for this example. There are three obvious ways in which one could use the given accuracies in our example to quantify feature importance: we could use the accuracy values themselves, we could use the normalized accuracy values so that the random noise neuron is measured as being of zero importance, and we could use the mutual information scores (as used for MIG). The follow code excerpt computes $D$ and $C$ from the disentanglement\_lib implementation for each of these three choices. In all three, the average decreases, when it should decrease.

\begin{lstlisting}
from dci import disentanglement, completeness
import numpy as np

def print_befores_and_afters(b,a):
    bd = disentanglement(b)['avg']
    ad = disentanglement(a)['avg']
    bc = completeness(b)['avg']
    ac = completeness(a)['avg']
    bavg = (bd+bc)/2
    aavg = (ad+ac)/2
    print(f'Before:\n{bd}, {bc}, {bavg}')
    print(f'After:\n{ad}, {ac}, {aavg}')

print('USING ACC AS IMPORTANCE')
b = np.array([[.75,.75],[.5,.5]])
a = np.array([[.75,.75],[.5,.7]])
print_befores_and_afters(b,a)

print('USING NORMED ACC AS IMPORTANCE')
b = np.array([[.5,.5],[.0,.0]])
a = np.array([[.5,.5],[.0,.2]])
print_befores_and_afters(b,a)

print('USING MUTUAL INFO AS IMPORTANCE')
b = np.array([[.1887,.1887],[.0,.0]])
a = np.array([[.1887,.1887],[.0,.1187]])
print_befores_and_afters(b,a)

\end{lstlisting}
\clearpage

\paragraph{SAP}
If we take the accuracy as roughly equal to the $R^2$ coefficient, then the SAP score for both factors before the change is 0.25, whereas after the change it is 0.25 for colour and 0.05 for shape, so the average decreases to 0.15.

\paragraph{SNC and NK.}
Our SNC metric is the average of the two chance-adjusted accuracies, $(0.5+0)/2 = 0.25$. NK is the drop in chance-adjusted accuracy after removing the aligned neuron, which is equal to 0.25 for colour and 0 for shape. In the variant where $z_1$ encodes shape to an accuracy of $70\%$, SNC becomes $(0.5+0.4)/2 = 0.45$, so correctly increases. NK is unchanged.

\section{Hinton Diagrams for Misaligned Factors}
Existing metrics simply assign each generative factor to the neuron that is most informative about it, as measured by mutual information or the weight in a linear classifier. Section \ref{sec:problems-with-existing} showed an example of this producing an incorrect alignment where multiple different factors are assigned to the same neuron. Our method, in contrast, enforces that all assignments are unique. Here we show some further examples of misalignments resulting from the method used by existing metrics. 

The following Hinton diagrams show the size of the square at $(i,j)$ is proportional to the mutual information between factor $j$ and neuron $i$.

\begin{figure}[hb]
    \centering
    \includegraphics[width=0.5\textwidth]{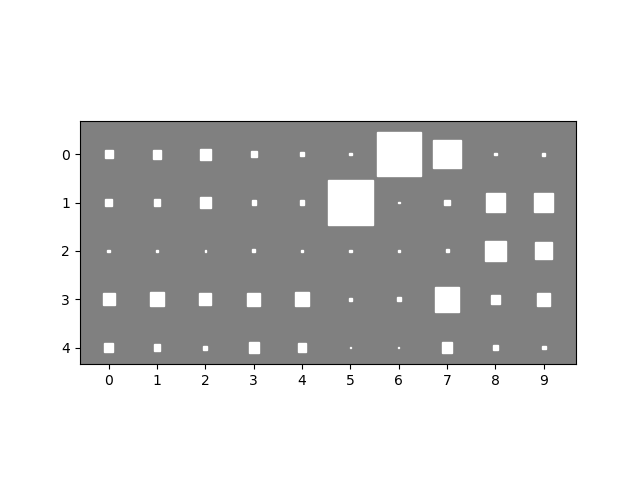}
    \caption{Hinton diagram showing alignment of factors (y-axis) to neurons (x-axis) for $\beta$-VAE on dsprites. Existing metrics incorrectly align both factor 3 and factor 4 to neuron 7, our method correctly aligns factor 3 to neuron 7 and factor 4 to neuron 3.}
    \label{fig:Hinton1}
\end{figure}

\begin{figure}
    \centering
    \includegraphics[width=0.5\textwidth]{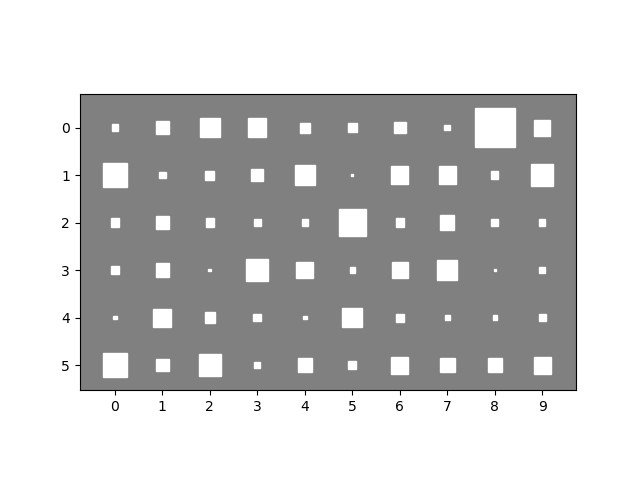}
    \caption{Hinton diagram showing alignment of factors (y-axis) to neurons (x-axis) for $\beta$-VAE on 3dshapes. Existing metrics incorrectly align both factor 2 and factor 4 to neuron 5, our method correctly aligns factor 2 to neuron 5 and factor 4 to neuron 1.}
    \label{fig:Hinton2}
\end{figure}

\begin{figure}
    \centering
    \includegraphics[width=0.5\textwidth]{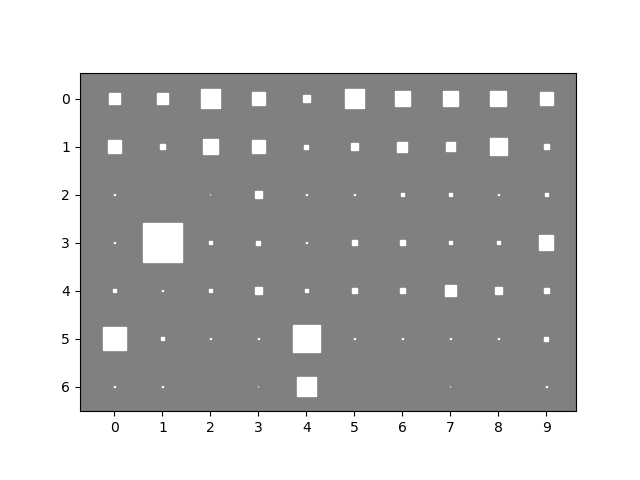}
    \caption{Hinton diagram showing alignment of factors (y-axis) to neurons (x-axis) for $\beta$-VAE on mpi3d. Existing metrics incorrectly align both factor 5 and factor 6 to neuron 4, our method correctly aligns factor 5 to neuron 0 and factor 6 to neuron 4.}
    \label{fig:Hinton3}
\end{figure}

\begin{figure}
    \centering
    \includegraphics[width=0.5\textwidth]{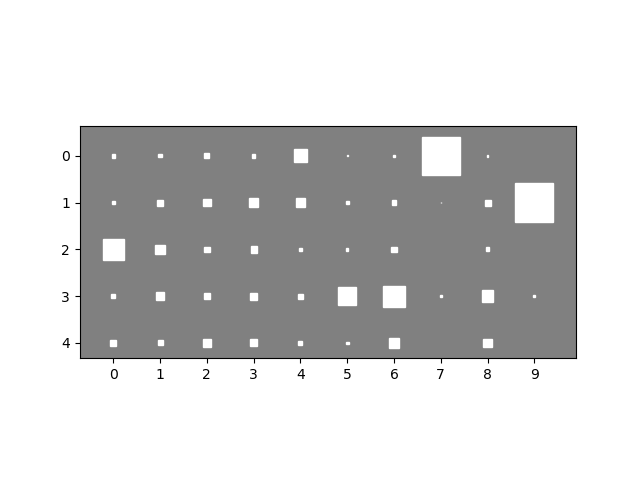}
    \caption{Hinton diagram showing alignment of factors (y-axis) to neurons (x-axis) for $\beta$-TCVAE on dsprites. Existing metrics incorrectly align both factor 3 and factor 4 to neuron 6, our method correctly aligns factor 3 to neuron 6 and factor 4 to neuron 8.}
    \label{fig:Hinton4}
\end{figure}

\begin{figure}
    \centering
    \includegraphics[width=0.5\textwidth]{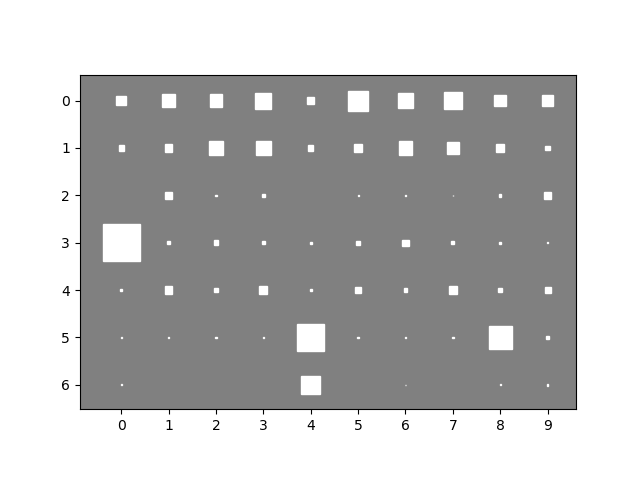}
    \caption{Hinton diagram showing alignment of factors (y-axis) to neurons (x-axis) for $\beta$-TCVAE on mpi3d. Existing metrics incorrectly align both factor 1 and factor 4 to neuron 3, our method correctly aligns factor 1 to neuron 3 and factor 4 to neuron 7.}
    \label{fig:Hinton5}
\end{figure}

\begin{figure}
    \centering
    \includegraphics[width=0.5\textwidth]{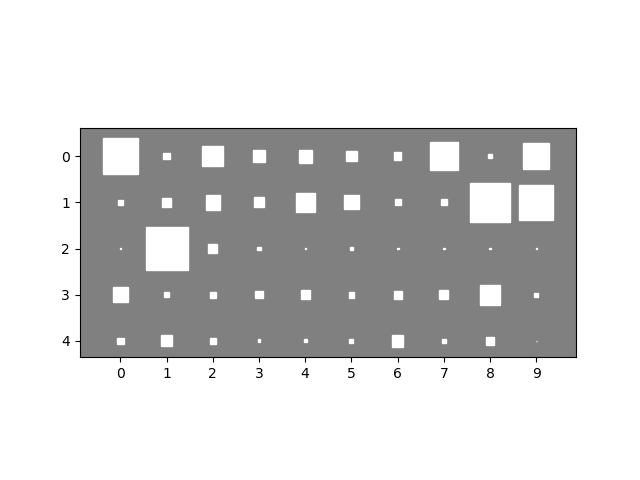}
    \caption{Hinton diagram showing alignment of factors (y-axis) to neurons (x-axis) for FactorVAE on dsprites. Existing metrics incorrectly align both factor 1 and factor 3 to neuron 8, our method correctly aligns factor 1 to neuron 8 and factor 3 to neuron 7.}
    \label{fig:Hinton6}
\end{figure}

\begin{figure}
    \centering
    \includegraphics[width=0.5\textwidth]{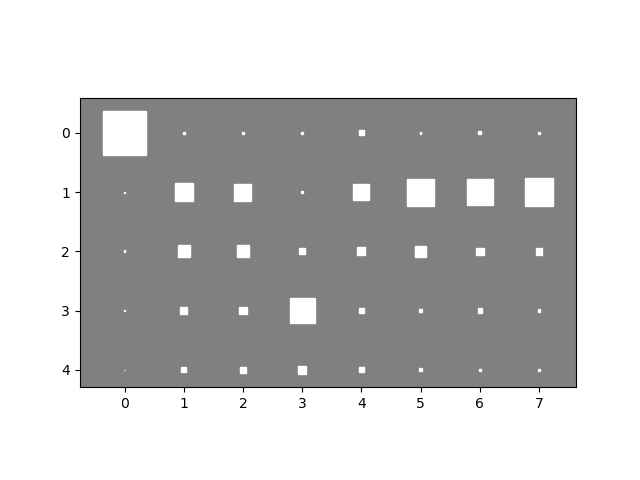}
    \caption{Hinton diagram showing alignment of factors (y-axis) to neurons (x-axis) for PartedVAE on dsprites. Existing metrics incorrectly align both factor 3 and factor 4 to neuron 3, our method correctly aligns factor 3 to neuron 3 and factor 4 to neuron 2.}
    \label{fig:Hinton7}
\end{figure}

\begin{figure}
    \centering
    \includegraphics[width=0.5\textwidth]{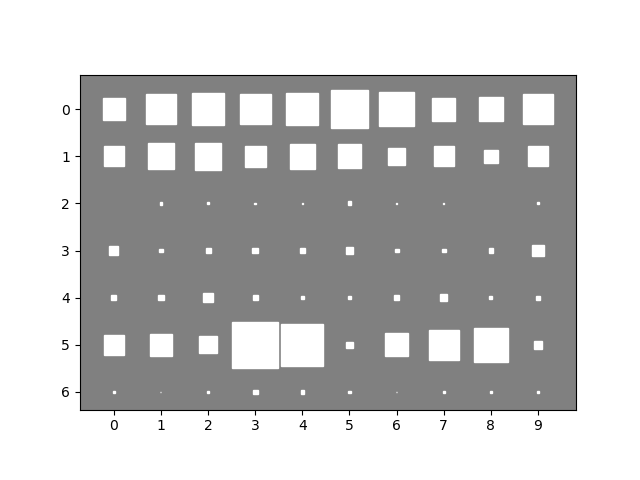}
    \caption{Hinton diagram showing alignment of factors (y-axis) to neurons (x-axis) for WeakDE on mpi3d. Existing metrics incorrectly align both factor 0 and factor 2 to neuron 5, our method correctly aligns factor 0 to neuron 5 and factor 2 to neuron 1.}
    \label{fig:Hinton8}
\end{figure}

\begin{figure}
    \centering
    \includegraphics[width=0.5\textwidth]{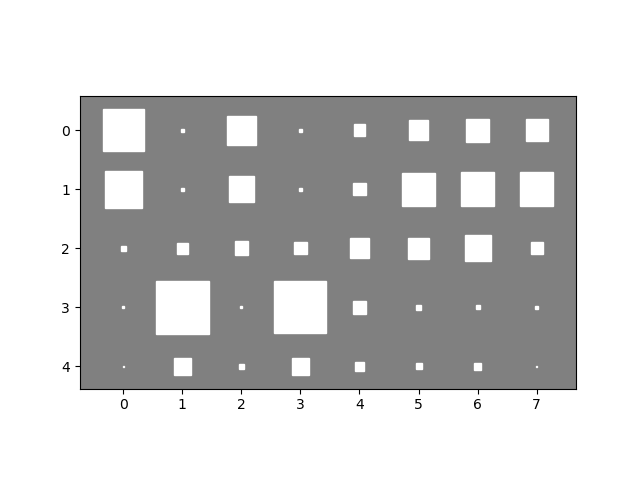}
    \caption{Hinton diagram showing alignment of factors (y-axis) to neurons (x-axis) for PartedVAE-semisupervised on dsprites. Existing metrics incorrectly align both factor 0 and factor 1 to neuron 0, our method correctly aligns factor 0 to neuron 0 and factor 1 to neuron 7.}
    \label{fig:Hinton9}
\end{figure}

\begin{figure}
    \centering
    \includegraphics[width=0.5\textwidth]{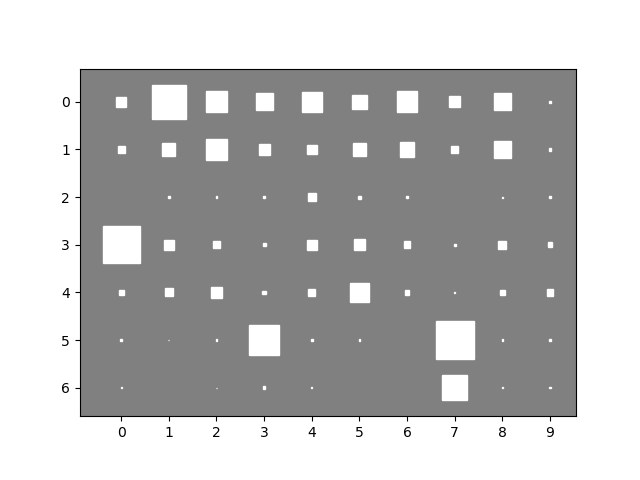}
    \caption{Hinton diagram showing alignment of factors (y-axis) to neurons (x-axis) for FactorVAE on mpi3d. Existing metrics incorrectly align both factor 5 and factor 6 to neuron 7, our method correctly aligns factor 5 to neuron 3 and factor 6 to neuron 7.}
    \label{fig:Hinton10}
\end{figure}

\begin{figure}
    \centering
    \includegraphics[width=0.5\textwidth]{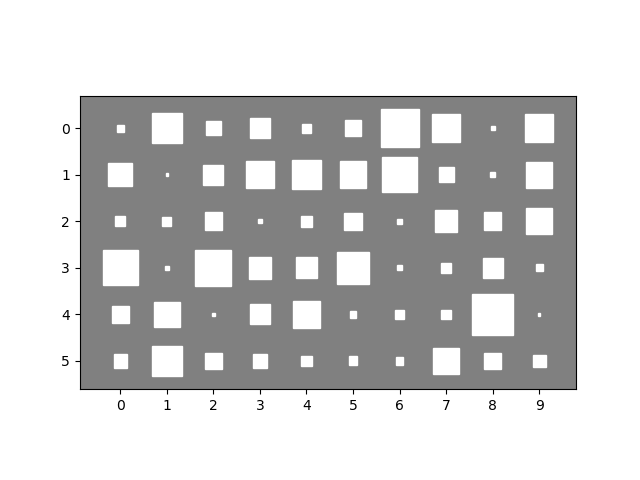}
    \caption{Hinton diagram showing alignment of factors (y-axis) to neurons (x-axis) for WeakDE on 3dshapes. Existing metrics incorrectly align both factor 0 and factor 1 to neuron 6, our method correctly aligns factor 0 to neuron 6 and factor 1 to neuron 4.}
    \label{fig:Hinton11}
\end{figure}

\clearpage

\section{Correlations of Different Versions of Existing Metrics with Compositional Generalization Performance}
As reported in Section \ref{subsec:correlation}, our metrics, restricted to the two novel feature types of size and shape, are more predictive of performance on compositional generalization than are existing metrics. When calculating the scores for other metrics, we took the average across all features. Table \ref{tab:full-correls} shows the correlation when restricting to just the features of shape and size. We show both the mean of these two features and the product (as for our metrics) of these two features. Note that the D component from DCI, and the IRS metric, are not computed feature-wise, so we cannot restrict it to just two features. These alternative variants of existing metrics perform better in some cases and worse in others. Overall, they are about equally predictive and, importantly, still all less predictive than our metrics, especially the SNC metric.

\begin{table}[hb]
\begin{tabular}{lrr}
\toprule
{} &  all datasets &  3dshapes \\
\midrule
SNC              &         \textbf{0.850} &     \textbf{0.849} \\
NK               &         \textit{0.716} &     0.710 \\
MIG              &         0.571 &     0.800 \\
MIG product of 2 &         0.570 &     0.640 \\
MIG mean of 2    &         0.514 &     0.786 \\
SAP              &         0.311 &     0.726 \\
SAP product of 2 &         0.626 &     0.544 \\
SAP mean of 2    &         0.453 &     0.625 \\
IRS              &         0.106 &     0.668 \\
D                &         0.457 &     0.472 \\
C                &         0.414 &     \textit{0.819} \\
C product of 2   &         0.288 &     0.681 \\
C mean of 2      &         0.280 &     0.770 \\
I                &         0.535 &     0.740 \\
I product of 2   &        -0.052 &     0.605 \\
I mean of 2      &         0.119 &     0.682 \\
DCI              &         0.493 &     0.717 \\
\bottomrule
\end{tabular}
\caption{Correlation (Pearson) of our metrics, and existing metrics, with accuracy on novel combinations. Best results in bold, second best italicized.}
\end{table} \label{tab:full-correls}
\clearpage

\section{Statistical Significance Calculation}
Here, we show the statistical significance of the correlations from Table \ref{tab:correls}. For the `all datasets'' experiment, there are 6 methods on 3 datasets, so 18 data points, giving t-value
\[
\frac{0.85 \sqrt{18-2}}{\sqrt{1-0.85^2}} \approx 6.45\,,
\]
which gives a p-value $<$ 1e-5.

For the '3dshapes' experiment, there are 6 methods on 1 datasets, so 6 data points, giving t-value
\[
\frac{0.85 \sqrt{6-2}}{\sqrt{1-0.85^2}} \approx 3.23\,,
\]
which gives a p-value $<$ 0.033, so still significant at $p<0.05$.

\clearpage
\section{Results of all metrics}
Table \ref{tab:all-other-metrics} shows the results of our metrics and existing metrics on the datasets and methods we test on, averaged over five runs for dsprites, eight for 3dshapes and six for mpi3d.

\begin{table*}[h]
\caption{Results of existing metrics on the datasets and methods we test on. The suffix `2p' and `2m' indicate, respectively, the product and mean across the two features being compositionally generalized, size and shape. When this suffix is absent, the figure is the mean across all features. Our own metrics, SNC and NK, are shown as the product across size and shape, because those are the figures used to calculate the correlation as reported in the main paper.} \label{tab:all-other-metrics}
\resizebox{1.1\textwidth}{!}{
\begin{tabular}{lrrrrrrrrrrrrrrrrrr}
\toprule
{} & \multicolumn{6}{l}{dsprites} & \multicolumn{6}{l}{3dshapes} & \multicolumn{6}{l}{mpi3d} \\
{} &    betaH & btcvae & factor &   pvae & pvae-ss & weakde &    betaH & btcvae & factor &    pvae & pvae-ss &  weakde &  betaH & btcvae & factor &   pvae & pvae-ss & weakde \\
\midrule
SNC   &    6.954 &  8.302 &  6.400 &  3.310 &   4.339 &  0.678 &    8.354 &  7.494 &  5.747 &  12.369 &  23.674 &  10.328 &  2.830 &  2.718 &  2.573 &  6.054 &   0.990 &  0.143 \\
NK    &    0.072 &  0.116 &  1.567 &  3.015 &  0.000 &  0.002 &    0.028 &  0.000 &  0.166 &  16.992 &  13.608 &   0.033 &  0.341 &  0.351 &  0.536 &  0.075 &   0.001 &  0.062 \\
MIG2p &    0.012 &  0.036 &  0.008 &  0.004 &   0.001 &  0.000 &    0.005 &  0.004 &  0.000 &   0.062 &   0.037 &   0.001 &  0.001 &  0.001 &  0.000 &  0.000 &   0.001 &  0.000 \\
MIG   &    0.085 &  0.144 &  0.088 &  0.070 &   0.058 &  0.014 &    0.063 &  0.062 &  0.032 &   0.279 &   0.255 &   0.026 &  0.104 &  0.171 &  0.086 &  0.086 &   0.023 &  0.013 \\
MIG2m &    0.130 &  0.204 &  0.177 &  0.068 &   0.047 &  0.014 &    0.069 &  0.051 &  0.018 &   0.265 &   0.236 &   0.038 &  0.023 &  0.030 &  0.012 &  0.072 &   0.046 &  0.005 \\
SAP   &    0.052 &  0.048 &  0.062 &  0.034 &   0.030 &  0.012 &    0.043 &  0.033 &  0.030 &   0.219 &   0.176 &   0.019 &  0.099 &  0.122 &  0.183 &  0.197 &   0.094 &  0.012 \\
SAP2p &    0.006 &  0.004 &  0.009 &  0.004 &   0.003 &  0.001 &    0.003 &  0.000 &  0.002 &   0.070 &   0.034 &   0.001 &  0.001 &  0.001 &  0.002 &  0.000 &   0.000 &  0.000 \\
SAP2m &    0.106 &  0.098 &  0.140 &  0.054 &   0.050 &  0.024 &    0.056 &  0.020 &  0.047 &   0.200 &   0.151 &   0.038 &  0.025 &  0.031 &  0.038 &  0.040 &   0.090 &  0.012 \\
IRS   &    0.448 &  0.601 &  0.558 &  0.610 &   0.732 &  0.491 &    0.385 &  0.490 &  0.510 &   0.720 &   0.708 &   0.466 &  0.427 &  0.604 &  0.545 &  0.739 &   0.622 &  0.531 \\
C2p   &    0.349 &  0.472 &  0.521 &  0.264 &   0.739 &  0.047 &    0.232 &  0.455 &  0.141 &   0.703 &   0.448 &   0.131 &  0.279 &  0.341 &  0.326 &  0.199 &   0.252 &  0.040 \\
C2m   &    0.576 &  0.686 &  0.750 &  0.505 &   0.395 &  0.219 &    0.494 &  0.505 &  0.387 &   0.842 &   0.714 &   0.369 &  0.554 &  0.551 &  0.545 &  0.859 &   0.503 &  0.230 \\
I2p   &    0.544 &  0.644 &  0.533 &  0.319 &   0.739 &  0.175 &    0.149 &  0.419 &  0.114 &   0.515 &   0.322 &   0.155 &  0.102 &  0.265 &  0.213 &  0.157 &   0.539 &  0.164 \\
I2m   &    0.745 &  0.783 &  0.733 &  0.580 &   0.503 &  0.424 &    0.387 &  0.395 &  0.365 &   0.826 &   0.698 &   0.407 &  0.327 &  0.372 &  0.305 &  0.508 &   0.840 &  0.442 \\
D     &    0.574 &  0.721 &  0.483 &  0.474 &   0.400 &  0.189 &    0.415 &  0.454 &  0.488 &   0.841 &   0.769 &   0.335 &  0.515 &  0.546 &  0.473 &  0.472 &   0.374 &  0.257 \\
C     &    0.544 &  0.595 &  0.453 &  0.461 &   0.459 &  0.084 &    0.320 &  0.372 &  0.296 &   0.792 &   0.687 &   0.219 &  0.457 &  0.477 &  0.437 &  0.331 &   0.231 &  0.134 \\
I     &    0.591 &  0.629 &  0.481 &  0.492 &   0.491 &  0.095 &    0.342 &  0.394 &  0.308 &   0.801 &   0.720 &   0.228 &  0.476 &  0.483 &  0.439 &  0.339 &   0.243 &  0.139 \\
DCI   &    0.012 &  0.040 &  0.009 &  0.005 &   0.000 &  0.000 &    0.003 &  0.001 &  0.000 &   0.075 &   0.039 &   0.001 &  0.001 &  0.001 &  0.000 &  0.000 &   0.002 &  0.000 \\
MED   &    0.068 &  0.113 &  0.071 &  0.050 &   0.042 &  0.008 &    0.048 &  0.044 &  0.029 &   0.231 &   0.213 &   0.026 &  0.071 &  0.147 &  0.076 &  0.079 &   0.020 &  0.013 \\
\bottomrule
\end{tabular}
}
\end{table*}
\section{Full Results}
Each of the following tables shows all results for a particular dataset and method combination. That is, each table shows results for single-neuron classification (SNC), neuron knockout (NK1 and NK2) and recognition of novel combinations of familiar features (NCFF) under the various settings described in the main paper.

For the compositional generalization settings, we indicate the values of two features excluded. Recall that we always exclude a combination of size and shape. So, for example ``NC 3-2'' means that the disentanglement model and the classifer trained on top of it, used a train set that excluded exactly those data points with size 3 and shape 2 (under some arbitrary ordering of the values of shape, e.g. 0=square, 1=circle, 2=crescent). 

For all settings, we report the results for all features (where measured). The feature lists for each data set are as follows:
\begin{itemize}
    \item \textbf{dsprites:} x-position (x), y-position (y), object size (size), object orientation (orient), and object shape (shape)
    \item \textbf{3dshapes:} floor colour (floor h), wall colour (wall h), object size (size), camera azimuth (orient), object shape (shape), and object colour (object h)
    \item \textbf{mpi3d:} azimuth of robot arm (hor), altitude of robot arm (vert), size of object (size), colour of object (obj h), shape of object (shape), height of camera above the object (cam he), and background colour (bg h)
\end{itemize}
We also report the accuracy on both of the novel features, i.e. the fraction of the points for which the classifier correctly predicted both size and shape. This is shown in the ``NC'' column.

\begin{table*}
\centering
\caption{Full results of $\beta$-VAE on dsprites for the main and normal test set settings.}
\begin{tabular}{llrrrrrr}
\toprule
                 &     &     x &     y &  size &  orient &  shape &    NC \\
\midrule
Normal Test set 0 & SNC & 40.78 & 46.84 & 44.80 &    7.34 &  52.17 & 23.17 \\
                 & linear & 70.11 & 72.09 & 48.02 &    7.24 &  44.31 & 17.55 \\
                 & NK1 & 17.59 & 20.80 & 48.83 &   50.04 &  98.87 &     - \\
                 & MLP & 89.63 & 92.87 & 99.04 &   65.54 &  99.64 & 98.83 \\
Normal Test set 1 & SNC & 29.74 & 32.44 & 45.90 &    7.17 &  49.99 & 23.52 \\
                 & linear & 57.68 & 62.44 & 50.26 &    9.12 &  46.75 & 20.06 \\
                 & NK1 & 17.43 & 21.17 & 48.97 &   50.10 &  99.17 &     - \\
                 & MLP & 91.22 & 92.46 & 99.22 &   66.79 &  99.53 & 98.86 \\
Normal Test set 2 & SNC & 31.58 & 28.61 & 32.82 &    8.02 &  47.69 & 16.25 \\
                 & linear & 61.54 & 62.28 & 42.46 &    7.33 &  41.20 & 16.44 \\
                 & NK1 & 17.26 & 18.61 & 74.60 &   45.49 &  99.21 &     - \\
                 & MLP & 90.43 & 91.39 & 98.92 &   65.73 &  99.78 & 98.79 \\
Normal Test set 3 & SNC & 31.49 & 28.75 & 45.15 &    6.61 &  48.19 & 25.62 \\
                 & linear & 68.12 & 71.43 & 44.93 &    7.99 &  53.61 & 24.95 \\
                 & NK1 & 23.67 & 23.21 & 47.74 &   46.00 &  99.46 &     - \\
                 & MLP & 94.07 & 94.23 & 98.88 &   67.51 &  99.87 & 98.78 \\
Normal Test set 4 & SNC & 38.37 & 43.16 & 31.20 &    7.49 &  51.50 & 16.36 \\
                 & linear & 59.68 & 57.23 & 50.29 &    6.09 &  53.58 & 20.61 \\
                 & NK1 & 20.02 & 19.29 & 80.06 &   52.03 &  99.52 &     - \\
                 & MLP & 89.92 & 89.44 & 99.02 &   61.74 &  99.07 & 98.38 \\
\bottomrule
\end{tabular}
\end{table*}

\begin{table*}
\centering
\caption{Full results of $\beta$-TCVAE on dsprites for the main and normal test set settings.}
\begin{tabular}{llrrrrrr}
\toprule
                  &     &     x &     y &  size &  orient &  shape &    zs \\
\midrule
normal test set 0 & SNC & 76.85 & 75.00 & 44.44 &   13.89 &  52.67 & 23.92 \\
                  & linear & 47.59 & 49.17 & 54.77 &   11.84 &  84.57 & 48.70 \\
                  & NK1 & 23.98 & 28.35 & 69.64 &   23.71 &  99.05 &     - \\
                  & MLP & 80.65 & 79.30 & 98.92 &   43.05 &  98.94 & 98.48 \\
normal test set 1 & SNC & 75.33 & 74.54 & 44.47 &   12.02 &  52.51 & 22.30 \\
                  & linear & 59.16 & 60.37 & 46.89 &   14.22 &  73.57 & 33.34 \\
                  & NK1 & 26.26 & 27.84 & 84.47 &   50.73 &  99.99 &     - \\
                  & MLP & 81.46 & 85.46 & 98.73 &   80.30 &  99.99 & 98.73 \\
normal test set 2 & SNC & 58.51 & 65.02 & 44.59 &   10.37 &  49.79 & 18.20 \\
                  & linear & 73.12 & 55.32 & 44.70 &   10.66 &  65.07 & 31.50 \\
                  & NK1 & 36.64 & 36.20 & 73.92 &   29.06 &  94.66 &     - \\
                  & MLP & 79.52 & 81.48 & 98.99 &   60.69 &  98.93 & 98.59 \\
normal test set 3 & SNC & 74.07 & 76.21 & 44.40 &    9.30 &  45.48 & 19.92 \\
                  & linear & 59.95 & 51.65 & 45.22 &    8.39 &  68.82 & 31.01 \\
                  & NK1 & 24.42 & 35.19 & 79.75 &   53.27 &  99.24 &     - \\
                  & MLP & 82.82 & 90.03 & 99.31 &   76.45 &  99.97 & 99.29 \\
normal test set 4 & SNC & 76.66 & 51.64 & 44.21 &   12.28 &  49.30 & 20.82 \\
                  & linear & 59.30 & 75.21 & 43.98 &    9.03 &  66.13 & 28.84 \\
                  & NK1 & 23.53 & 38.40 & 80.50 &   33.70 &  99.96 &     - \\
                  & MLP & 81.54 & 87.05 & 98.97 &   72.86 &  99.96 & 98.95 \\
\bottomrule
\end{tabular}
\end{table*}

\begin{table*}
\centering
\caption{Full results of FactorVAE on dsprites for the main and normal test set settings.}
\begin{tabular}{llrrrrrr}
\toprule
                 &     &     x &     y &  size &  orient &  shape &    NC \\
\midrule
Normal Test set 0 & SNC & 11.28 & 17.09 & 45.07 &    4.49 &  43.69 & 18.85 \\
                 & linear & 31.39 & 29.82 & 49.05 &    4.66 &  46.28 & 20.45 \\
                 & NK1 & 41.93 & 22.33 & 33.99 &   19.36 &  93.18 &     - \\
                 & MLP & 76.29 & 77.66 & 97.36 &   33.98 &  98.58 & 96.29 \\
Normal Test set 1 & SNC & 10.81 &  7.93 & 46.25 &    3.96 &  46.74 & 22.18 \\
                 & linear & 20.26 & 21.70 & 49.82 &    3.89 &  43.93 & 22.90 \\
                 & NK1 & 22.60 & 40.56 & 35.44 &   18.87 &  93.65 &     - \\
                 & MLP & 74.98 & 74.87 & 96.95 &   32.41 &  97.45 & 95.15 \\
Normal Test set 2 & SNC & 11.81 & 17.05 & 47.09 &    4.13 &  43.17 & 20.49 \\
                 & linear &  5.53 & 27.45 & 49.76 &    6.27 &  53.39 & 27.48 \\
                 & NK1 & 44.08 & 42.64 & 39.35 &   23.61 &  94.90 &     - \\
                 & MLP & 78.16 & 78.72 & 98.43 &   42.24 &  98.96 & 97.72 \\
Normal Test set 3 & SNC & 16.38 & 17.62 & 48.49 &    5.78 &  47.36 & 22.66 \\
                 & linear & 31.39 & 32.27 & 52.65 &    3.97 &  46.45 & 24.15 \\
                 & NK1 & 25.68 & 20.32 & 34.40 &   18.53 &  95.26 &     - \\
                 & MLP & 76.09 & 77.75 & 97.86 &   36.38 &  98.55 & 96.82 \\
Normal Test set 4 & SNC &  7.60 &  9.13 & 48.08 &    3.38 &  44.32 & 21.36 \\
                 & linear & 29.74 & 25.99 & 51.36 &    4.30 &  48.41 & 27.40 \\
                 & NK1 & 44.79 & 46.11 & 37.80 &   23.25 &  92.10 &     - \\
                 & MLP & 74.00 & 75.37 & 97.00 &   37.73 &  98.55 & 96.03 \\
\bottomrule
\end{tabular}
\end{table*}

\begin{table*}
\centering
\caption{Full results of PartedVAE on dsprites for the main and normal test set settings.}
\begin{tabular}{llrrrrrr}
\toprule
                 &     &     x &     y &  size &  orient &  shape &    NC \\
\midrule
Normal Test set 0 & SNC & 47.10 & 12.54 & 27.42 &    9.11 &  43.62 & 13.83 \\
                 & linear & 42.23 & 48.93 & 42.88 &    8.64 &  52.07 & 22.91 \\
                 & NK1 &  5.90 & 56.56 & 48.59 &    7.91 &  75.46 &     - \\
                 & MLP & 72.14 & 68.87 & 75.58 &   32.03 &  95.87 & 73.99 \\
Normal Test set 1 & SNC & 13.76 &  8.47 & 29.44 &    7.54 &  40.12 & 11.44 \\
                 & linear & 31.00 & 36.90 & 33.00 &    7.50 &  49.71 & 15.37 \\
                 & NK1 & 10.05 & 37.85 & 67.24 &    9.31 &  80.02 &     - \\
                 & MLP & 72.75 & 74.11 & 57.29 &   29.83 &  97.91 & 56.84 \\
Normal Test set 2 & SNC & 10.13 &  6.04 & 33.26 &    8.11 &  53.70 & 16.78 \\
                 & linear & 30.84 & 31.57 & 39.80 &   11.07 &  60.01 & 23.97 \\
                 & NK1 &  9.97 &  5.95 & 33.05 &    8.07 &  54.04 & 16.67 \\
                 & MLP & 70.26 & 74.17 & 78.93 &   55.02 &  98.96 & 78.51 \\
Normal Test set 3 & SNC & 13.31 & 48.36 & 43.50 &    4.87 &  38.62 & 17.35 \\
                 & linear & 52.68 & 41.83 & 42.51 &    4.26 &  46.01 & 17.42 \\
                 & NK1 & 56.76 &  5.44 & 43.97 &    7.50 &  85.22 &     - \\
                 & MLP & 66.55 & 75.06 & 74.94 &   20.91 &  94.07 & 72.75 \\
Normal Test set 4 & SNC & 42.76 & 15.96 & 31.59 &    8.10 &  46.75 & 14.95 \\
                 & linear & 37.48 & 46.10 & 46.31 &   11.65 &  45.52 & 20.31 \\
                 & NK1 &  6.37 & 29.60 & 59.92 &    9.25 &  80.02 &     - \\
                 & MLP & 70.27 & 73.46 & 88.22 &   50.11 &  98.99 & 87.70 \\
\bottomrule
\end{tabular}
\end{table*}

\begin{table*}
\centering
\caption{Full results of PartedVAE-semisupervised on dsprites for the main and normal test set settings.}
\begin{tabular}{llrrrrrr}
\toprule
                 &     &     x &     y &  size &  orient &  shape &    NC \\
\midrule
Normal Test set 0 & SNC & 19.43 & 39.12 & 46.53 &    5.94 &  39.56 &     - \\
                 & linear & 40.02 & 36.01 & 47.00 &    3.24 &  43.69 & 20.85 \\
                 & NK1 & 23.91 & 10.08 & 42.01 &    4.63 &  72.54 &     - \\
                 & MLP & 71.62 & 71.87 & 68.06 &    3.58 &  64.98 & 50.67 \\
Normal Test set 1 & SNC & 19.43 & 39.12 & 46.53 &    5.94 &  39.56 & 18.11 \\
                 & linear &  3.11 &  3.11 & 16.78 &    2.48 &  33.27 &  5.56 \\
                 & NK1 & 23.91 & 10.08 & 42.01 &    4.63 &  72.54 &     - \\
                 & MLP &  3.14 &  3.11 & 16.52 &    2.46 &  33.39 &  5.43 \\
Normal Test set 2 & SNC & 19.43 & 39.12 & 46.53 &    5.94 &  39.56 & 18.11 \\
                 & linear & 34.37 & 38.18 & 44.65 &    6.63 &  46.61 & 23.49 \\
                 & NK1 &  9.70 & 47.66 & 47.17 &    7.46 &  72.25 &     - \\
                 & MLP & 63.27 & 70.20 & 61.73 &   13.32 &  88.56 & 58.16 \\
Normal Test set 3 & SNC &  7.94 &  4.48 & 29.43 &    9.32 &  48.59 & 14.17 \\
                 & linear &  7.86 &  4.40 & 29.43 &    9.52 &  48.59 & 14.17 \\
                 & NK1 &  9.70 & 47.66 & 47.17 &    7.46 &  72.25 &     - \\
                 & MLP & 16.77 & 67.65 & 57.61 &   15.13 &  81.59 &     - \\
Normal Test set 4 & SNC & 43.74 & 21.89 & 37.34 &    7.95 &  48.38 & 15.66 \\
                 & linear & 43.56 & 22.18 & 37.34 &    7.89 &  48.38 & 15.66 \\
                 & NK1 &  5.48 & 30.74 & 51.88 &    5.21 &  80.36 &     - \\
                 & MLP & 71.56 & 72.21 & 76.74 &   26.32 &  90.54 & 72.10 \\
\bottomrule
\end{tabular}
\end{table*}

\begin{table*}
\centering
\caption{Full results of WeakDE on dsprites for the main and normal test set settings.}
\begin{tabular}{llrrrrrr}
\toprule
                 &     &     x &     y &  size &  orient &  shape &    NC \\
\midrule
Normal Test set 0 & SNC &  7.51 &  8.11 & 29.11 &    3.04 &  37.47 & 11.79 \\
                 & linear & 37.30 & 42.80 & 48.32 &    3.51 &  42.66 & 18.55 \\
                 & NK1 & 71.77 & 67.55 & 77.71 &   39.14 &  98.89 &     - \\
                 & MLP & 81.40 & 82.05 & 92.76 &   46.80 &  99.07 & 92.13 \\
Normal Test set 1 & SNC &  8.14 &  8.15 & 22.10 &    3.07 &  37.62 &  8.61 \\
                 & linear & 39.39 & 38.30 & 28.17 &    3.53 &  40.74 & 10.64 \\
                 & NK1 & 66.54 & 66.77 & 79.80 &   39.02 &  98.79 &     - \\
                 & MLP & 80.93 & 80.02 & 87.71 &   44.05 &  98.93 & 87.01 \\
Normal Test set 2 & SNC &  8.35 &  7.53 & 24.05 &    3.25 &  37.67 &  9.12 \\
                 & linear & 41.35 & 41.21 & 44.33 &    3.47 &  42.61 & 18.22 \\
                 & NK1 & 64.68 & 71.90 & 83.36 &   40.52 &  98.92 &     - \\
                 & MLP & 82.55 & 82.89 & 94.37 &   47.91 &  98.73 & 93.49 \\
Normal Test set 3 & SNC &  8.61 &  7.80 & 26.70 &    3.22 &  38.92 & 10.73 \\
                 & linear & 38.94 & 38.46 & 43.26 &    3.87 &  43.54 & 17.74 \\
                 & NK1 & 67.53 & 71.59 & 80.87 &   40.75 &  98.92 &     - \\
                 & MLP & 82.14 & 83.36 & 92.53 &   48.07 &  98.97 & 91.87 \\
Normal Test set 4 & SNC &  7.28 &  6.28 & 21.78 &    3.18 &  38.30 &  8.30 \\
                 & linear & 38.70 & 41.32 & 35.41 &    3.10 &  42.46 & 13.88 \\
                 & NK1 & 71.17 & 73.76 & 81.51 &   39.35 &  98.89 &     - \\
                 & MLP & 81.78 & 82.90 & 89.26 &   47.07 &  98.87 & 88.51 \\
\bottomrule
\end{tabular}
\end{table*}

\begin{table*}
\centering
\caption{Full results of $\beta$-VAE on 3dshapes for the main and normal test set settings.}
\begin{tabular}{llrrrrrrr}
\toprule
                 &     &  floor h &  wall h &   size &  orient &  shape &  obj h &     NC \\
\midrule
Normal Test set 0 & SNC &    20.23 &   22.98 &  35.49 &   24.63 &  36.73 &  20.61 &  12.85 \\
                 & linear &    99.06 &   99.25 &  55.55 &   87.31 &  35.56 &  99.03 &  13.93 \\
                 & NK1 &    98.92 &   97.14 &  87.62 &   42.19 &  99.91 &  99.10 &      - \\
                 & MLP &   100.00 &  100.00 & 100.00 &   99.92 & 100.00 & 100.00 & 100.00 \\
Normal Test set 1 & SNC &    21.86 &   16.99 &  16.37 &   15.77 &  41.45 &  17.30 &   7.11 \\
                 & linear &    99.00 &   99.12 &  15.72 &   96.39 &  64.16 &  99.05 &   8.39 \\
                 & NK1 &    96.53 &   98.18 &  94.16 &   70.73 &  98.91 &  99.00 &      - \\
\bottomrule
\end{tabular}
\end{table*}

\begin{table*}
\centering
\caption{Full results of $\beta$-TCVAE on 3dshapes for the main and normal test set settings.}
\begin{tabular}{llrrrrrrr}
\toprule
                 &     &  floor h &  wall h &   size &  orient &  shape &  obj h &     NC \\
\midrule
Normal Test set 0 & SNC &    31.69 &   20.30 &  36.03 &   20.79 &  39.68 &  21.05 &  14.58 \\
                 & linear &    99.13 &   99.08 &  59.18 &   82.33 &  29.42 &  99.03 &  17.38 \\
                 & NK1 &    85.91 &   95.27 &  92.73 &   65.54 &  99.37 &  97.91 &      - \\
                 & MLP &   100.00 &  100.00 &  98.76 &   99.98 &  99.81 & 100.00 &  98.51 \\
Normal Test set 1 & SNC &    27.81 &   19.87 &  21.44 &   25.79 &  39.33 &  25.51 &   9.74 \\
                 & linear &    98.83 &   98.75 &  31.02 &   83.30 &  37.69 &  98.96 &  10.72 \\
                 & NK1 &    88.46 &   98.28 &  92.15 &   75.89 &  99.19 &  99.40 &      - \\
\bottomrule
\end{tabular}
\end{table*}

\begin{table*}
\centering
\caption{Full results of FactorVAE on 3dshapes for the main and normal test set settings.}
\begin{tabular}{llrrrrrrr}
\toprule
                 &     &  floor h &  wall h &  size &  orient &  shape &  obj h &    NC \\
\midrule
Normal Test set 0 & SNC &    30.63 &   23.69 & 31.55 &   15.74 &  31.90 &  12.77 & 10.63 \\
                 & linear &    98.94 &   98.97 & 38.43 &   38.27 &  31.99 &  74.98 & 11.52 \\
                 & NK1 &    83.37 &   89.94 & 79.34 &   67.52 &  98.39 &  94.20 &     - \\
                 & MLP &   100.00 &   99.54 & 67.03 &   98.46 &  98.90 &  98.99 & 66.20 \\
Normal Test set 1 & SNC &    23.26 &   14.53 & 24.64 &   10.35 &  33.13 &  17.75 &  8.01 \\
                 & linear &    96.79 &   99.02 & 18.76 &   57.48 &  54.74 &  86.12 &  9.14 \\
                 & NK1 &    87.50 &   91.06 & 94.80 &   64.25 &  99.03 &  94.86 &     - \\
\bottomrule
\end{tabular}
\end{table*}

\begin{table*}
\centering
\caption{Full results of PartedVAE on 3dshapes for the main and normal test set settings.}
\begin{tabular}{llrrrrrrr}
\toprule
                 &     &  floor h &  wall h &  size &  orient &  shape &  obj h &    NC \\
\midrule
Normal Test set 0 & SNC &    99.86 &   37.53 & 38.02 &   99.53 &  93.45 &  82.14 & 35.10 \\
                 & linear &    89.72 &   96.97 & 17.46 &   78.20 &  82.05 &  82.20 & 13.72 \\
                 & NK1 &    26.44 &   94.22 & 18.01 &    8.04 &  13.02 &  61.88 &     - \\
                 & MLP &    99.07 &   98.86 & 90.97 &   99.07 &  99.03 &  98.94 & 90.13 \\
Normal Test set 1 & SNC &     7.50 &    6.68 & 30.74 &    8.29 &  44.40 &  13.48 &     - \\
                 & linear &    99.04 &   90.69 & 33.07 &   15.41 &  89.03 &  98.97 & 30.04 \\
                 & NK1 &    14.24 &   90.43 & 20.57 &    9.57 &  17.77 &  56.55 &     - \\
\bottomrule
\end{tabular}
\end{table*}

\begin{table*}
\centering
\caption{Full results of PartedVAE-semisupervised on 3dshapes for the main and normal test set settings.}
\begin{tabular}{llrrrrrrr}
\toprule
                 &     &  floor h &  wall h &  size &  orient &  shape &  obj h &    NC \\
\midrule
Normal Test set 0 & SNC &    90.16 &   99.79 & 27.72 &   87.55 &  96.70 &  43.88 & 25.92 \\
                 & linear &    61.00 &   62.00 & 41.00 &    8.00 &  99.00 &  51.00 & 40.00 \\
                 & NK1 &    99.98 &   15.74 & 28.94 &   15.21 &  21.65 &  51.11 &     - \\
                 & MLP &    99.97 &   99.33 & 79.00 &   94.17 &  93.03 &  98.61 & 75.68 \\
Normal Test set 1 & SNC &    99.89 &   99.78 & 68.99 &   18.86 &  94.32 &  95.50 & 64.38 \\
                 & linear &    81.55 &   86.23 & 82.90 &   44.33 &  98.64 &  78.13 & 82.02 \\
                 & NK1 &    26.95 &   35.40 & 46.68 &   47.05 &  81.65 &  38.12 &     - \\
\bottomrule
\end{tabular}
\end{table*}

\begin{table*}
\centering
\caption{Full results of WeakDE on 3dshapes for the main and normal test set settings.}
\begin{tabular}{llrrrrrrr}
\toprule
                 &     &  floor h &  wall h &  size &  orient &  shape &  obj h &    NC \\
\midrule
Normal Test set 0 & SNC &    19.50 &   22.68 & 17.03 &    9.62 &  27.94 &  19.02 &  4.98 \\
                 & linear &    99.22 &   99.31 & 13.61 &   15.92 &  35.06 &  99.02 &  4.77 \\
                 & NK1 &    98.97 &   99.05 & 88.36 &   98.51 &  99.02 &  98.87 &     - \\
                 & MLP &   100.00 &  100.00 & 99.92 &   99.18 & 100.00 & 100.00 & 99.92 \\
Normal Test set 1 & SNC &    22.14 &   23.58 & 17.98 &    9.43 &  50.12 &  19.52 &  9.04 \\
                 & linear &    98.80 &   98.96 & 32.86 &   23.15 &  70.65 &  96.92 & 23.18 \\
                 & NK1 &    98.25 &   98.94 & 95.03 &   91.48 &  99.09 &  98.47 &     - \\
\bottomrule
\end{tabular}
\end{table*}

\begin{table*}
\centering
\caption{Full results of $\beta$-VAE on mpi3d for the main and normal test set settings.}
\begin{tabular}{llrrrrrrrr}
\toprule
                 &     &   hor &  vert &  size &  obj h &  shape &  cam he &   bg h &    NC \\
\midrule
Normal Test set 0 & SNC &  5.30 &  5.24 & 66.06 &  73.80 &  26.15 &   94.84 &  54.61 & 17.39 \\
                 & linear & 17.63 & 11.68 & 77.79 &  76.17 &  36.94 &   99.71 &  48.30 & 28.21 \\
                 & NK1 & 66.18 & 40.06 & 92.69 &  70.65 &  74.81 &   88.90 &  50.87 &     - \\
                 & MLP & 86.51 & 70.23 & 98.92 &  99.82 &  88.28 &  100.00 &  99.99 & 87.61 \\
Normal Test set 1 & SNC &  6.26 &  4.45 & 66.69 &  78.82 &  27.08 &   74.31 &  52.44 & 18.36 \\
                 & linear & 16.87 &  9.24 & 76.29 &  88.91 &  34.92 &  100.00 &  45.30 & 25.93 \\
                 & NK1 & 59.51 & 40.68 & 91.87 &  69.46 &  71.61 &   91.04 &  70.78 &     - \\
\bottomrule
\end{tabular}
\end{table*}

\begin{table*}
\centering
\caption{Full results of $\beta$-TCVAE on mpi3d for the main and normal test set settings.}
\begin{tabular}{llrrrrrrrr}
\toprule
                 &     &   hor &  vert &  size &  obj h &  shape &  cam he &   bg h &    NC \\
\midrule
Normal Test set 0 & SNC &  5.13 &  4.14 & 65.64 &  70.81 &  23.31 &   67.86 &  55.50 & 14.34 \\
                 & linear & 13.54 & 10.34 & 75.53 &  70.25 &  32.45 &   99.84 &  40.92 & 23.82 \\
                 & NK1 & 56.75 & 40.24 & 92.83 &  80.80 &  66.90 &   91.88 &  46.96 &     - \\
                 & MLP & 85.59 & 66.92 & 98.79 &  99.44 &  86.87 &  100.00 & 100.00 & 86.06 \\
Normal Test set 1 & SNC &  5.91 &  5.60 & 69.62 &  71.87 &  27.90 &   67.77 &  55.60 & 18.94 \\
                 & linear & 17.20 & 12.70 & 76.87 &  75.50 &  35.48 &  100.00 &  46.89 & 26.01 \\
                 & NK1 & 56.99 & 40.43 & 92.79 &  80.58 &  66.93 &   91.78 &  46.76 &     - \\
\bottomrule
\end{tabular}
\end{table*}

\begin{table*}
\centering
\caption{Full results of FactorVAE on mpi3d for the main and normal test set settings.}
\begin{tabular}{llrrrrrrrr}
\toprule
                 &     &   hor &  vert &  size &  obj h &  shape &  cam he &  bg h &    NC \\
\midrule
Normal Test set 0 & SNC &  6.14 &  4.98 & 64.12 &  40.64 &  24.57 &   68.75 & 46.87 & 15.79 \\
                 & linear & 12.66 &  8.85 & 63.69 &  52.95 &  36.11 &   99.10 & 36.79 & 23.72 \\
                 & NK1 & 47.85 & 29.58 & 89.10 &  74.60 &  63.35 &   81.47 & 59.91 &     - \\
                 & MLP & 77.95 & 59.73 & 97.43 &  98.88 &  80.46 &  100.00 & 99.83 & 78.84 \\
Normal Test set 1 & SNC &  5.15 &  4.48 & 56.76 &  69.91 &  26.81 &   75.57 & 52.86 & 15.10 \\
                 & linear &  6.30 &  9.78 & 61.18 &  76.65 &  35.10 &   99.33 & 35.35 & 21.39 \\
                 & NK1 & 54.76 & 32.60 & 91.11 &  66.24 &  65.88 &   80.98 & 62.50 &     - \\
\bottomrule
\end{tabular}
\end{table*}

\begin{table*}
\centering
\caption{Full results of PartedVAE on mpi3d for the main and normal test set settings.}
\begin{tabular}{llrrrrrrrr}
\toprule
                 &     &   hor &  vert &  size &  obj h &  shape &  cam he &  bg h &    NC \\
\midrule
Normal Test set 0 & SNC & 10.13 &  6.90 & 59.61 &  20.70 &  85.20 &   79.62 & 21.03 & 50.95 \\
                 & linear & 13.61 &  5.03 & 74.11 &  30.09 & 100.00 &   99.84 & 28.68 & 19.89 \\
                 & NK1 & 18.90 &  8.51 & 74.45 &  76.52 & 100.00 &   92.26 & 32.14 &     - \\
                 & MLP & 38.36 & 26.28 & 81.37 &  93.46 & 100.00 &   99.97 & 37.50 & 31.20 \\
Normal Test set 1 & SNC &  9.03 &  6.75 & 59.86 &  23.50 &  99.46 &   89.06 & 21.09 & 59.42 \\
                 & linear & 13.81 &  4.39 & 72.60 &  30.92 & 100.00 &   98.54 & 28.38 & 19.38 \\
                 & NK1 & 16.40 &  6.67 & 76.40 &  71.57 & 100.00 &   94.37 & 31.05 &     - \\
\bottomrule
\end{tabular}
\end{table*}

\begin{table*}
\centering
\caption{Full results of PartedVAE-semisupervised on mpi3d for the main and normal test set settings.}
\begin{tabular}{llrrrrrrrr}
\toprule
                 &     &   hor &  vert &  size &  obj h &  shape &  cam he &  bg h &    NC \\
\midrule
Normal Test set 0 & SNC &  6.72 &  3.87 & 59.61 &  22.89 &  19.24 &   86.08 & 34.16 & 11.75 \\
                 & linear & 11.16 &  5.91 & 63.43 &  33.39 &  24.05 &   99.19 & 35.72 & 14.95 \\
                 & NK1 & 23.80 & 16.78 & 74.27 &  71.11 &  31.55 &  100.00 & 50.64 &     - \\
                 & MLP & 25.23 & 16.15 & 74.58 &  72.87 &  31.49 &  100.00 & 51.22 & 23.16 \\
Normal Test set 1 & SNC &  5.11 &  4.01 & 64.20 &  26.78 &  18.82 &   77.92 & 53.51 & 11.86 \\
                 & linear & 13.61 &  5.10 & 71.16 &  38.57 &  26.76 &   99.79 & 56.38 & 17.84 \\
                 & NK1 & 25.79 & 12.96 & 79.32 &  84.73 &  33.45 &  100.00 & 60.71 &     - \\
\bottomrule
\end{tabular}
\end{table*}

\begin{table*}
\centering
\caption{Full results of WeakDE on mpi3d for the main and normal test set settings.}
\begin{tabular}{llrrrrrrrr}
\toprule
                 &     &   hor &  vert &  size &  obj h &  shape &  cam he &   bg h &    NC \\
\midrule
Normal Test set 0 & SNC &  0.57 &  3.84 & 52.53 &  20.78 &  18.40 &   71.63 &  35.50 &  9.66 \\
                 & linear & 15.81 & 17.02 & 58.85 &  33.60 &  25.14 &  100.00 &  62.21 & 15.30 \\
                 & NK1 & 70.17 & 48.03 & 88.73 &  82.23 &  53.53 &  100.00 &  93.98 &     - \\
                 & MLP & 76.40 & 60.47 & 90.95 &  98.67 &  60.03 &  100.00 &  99.99 & 56.00 \\
Normal Test set 1 & SNC &  5.16 &  4.72 & 52.89 &  19.39 &  19.53 &   72.53 &  35.00 & 10.75 \\
                 & linear & 18.03 & 16.50 & 65.62 &  25.35 &  27.76 &  100.00 &  79.91 & 17.64 \\
                 & NK1 & 75.25 & 53.43 & 89.21 &  84.39 &  55.02 &  100.00 &  98.85 &     - \\
\bottomrule
\end{tabular}
\end{table*}

\begin{table*}
\centering
\caption{Full results of $\beta$-VAE on dsprites for the compositional generalization setting.}
\begin{tabular}{llrrrrrr}
\toprule
       &    &     x &     y &  size &  orient &  shape &   NC \\
\midrule
NC 0-0 & CG linear & 30.14 & 19.38 &  7.91 &    2.64 &   0.01 & 0.00 \\
       & CG & 95.77 & 93.95 &  7.03 &   10.84 &  78.52 & 5.08 \\
NC 5-2 & CG linear & 21.06 & 21.44 & 18.52 &    8.38 &   8.94 & 0.00 \\
       & CG & 44.42 & 51.18 &  0.01 &   76.80 &  96.88 & 0.00 \\
NC 2-1 & CG linear &  3.12 &  3.12 &  0.00 &    2.50 &   0.00 & 0.00 \\
       & CG & 89.07 & 87.75 &  5.55 &   38.73 &  99.69 & 5.49 \\
NC 3-1 & CG linear & 86.13 & 30.04 &  0.00 &    9.85 &   2.16 & 0.00 \\
       & CG & 98.44 & 97.87 &  8.42 &   73.54 &  99.94 & 8.37 \\
NC 3-2 & CG linear & 22.47 & 20.41 &  0.00 &    9.40 &   9.70 & 0.00 \\
       & CG & 79.55 & 84.32 &  0.00 &   92.19 &  99.39 & 0.00 \\
\bottomrule
\end{tabular}
\end{table*}

\begin{table*}
\centering
\caption{Full results of $\beta$-TCVAE on dsprites for the compositional generalization setting.}
\begin{tabular}{llrrrrrr}
\toprule
       &    &     x &     y &  size &  orient &  shape &   NC \\
\midrule
NC 0-0 & CG linear & 31.96 & 38.55 &  0.55 &    2.96 &   0.00 & 0.00 \\
       & CG & 91.24 & 96.66 &  0.00 &    6.40 &  19.02 & 0.00 \\
NC 5-2 & CG linear & 23.32 & 22.40 &  0.00 &   21.37 &   0.00 & 0.00 \\
       & CG & 22.06 & 19.77 &  0.00 &   40.17 &  95.48 & 0.00 \\
NC 2-1 & CG linear & 65.64 & 66.06 &  0.42 &   15.96 &  26.21 & 0.01 \\
       & CG & 97.68 & 95.48 &  0.00 &   78.81 &  99.99 & 0.00 \\
NC 3-1 & CG linear & 99.54 & 99.39 &  0.00 &   82.19 & 100.00 & 0.00 \\
       & CG & 99.54 & 99.39 &  0.00 &   82.19 & 100.00 & 0.00 \\
NC 3-2 & CG linear & 22.60 & 28.62 &  0.06 &    5.32 &  63.63 & 0.02 \\
       & CG & 38.94 & 51.74 &  0.00 &   94.16 & 100.00 & 0.00 \\
\bottomrule
\end{tabular}
\end{table*}

\begin{table*}
\centering
\caption{Full results of FactorVAE on dsprites for the compositional generalization setting.}
\begin{tabular}{llrrrrrr}
\toprule
       &    &     x &     y &  size &  orient &  shape &    NC \\
\midrule
NC 0-0 & CG linear & 37.29 & 29.55 &  2.13 &    2.91 &   0.00 &  0.00 \\
       & CG & 96.60 & 95.51 &  0.37 &    7.52 &  31.49 &  0.00 \\
NC 5-2 & CG linear & 33.84 & 35.84 &  0.00 &    7.23 &   6.25 &  0.00 \\
       & CG & 34.67 & 20.87 &  0.20 &   46.77 &  67.82 &  0.20 \\
NC 2-1 & CG linear & 26.98 & 27.80 & 11.16 &    7.22 &  57.42 &  1.57 \\
       & CG & 26.98 & 27.80 & 11.16 &    7.22 &  57.42 &  1.57 \\
NC 3-1 & CG linear & 24.69 & 17.07 &  2.32 &    4.78 &  31.77 &  0.00 \\
       & CG & 89.27 & 85.19 & 30.59 &   28.76 &  98.82 & 30.11 \\
NC 3-2 & CG linear & 23.10 & 17.11 &  0.00 &    6.23 &  46.23 &  0.00 \\
       & CG & 41.77 & 39.57 &  0.00 &   64.69 &  97.54 &  0.00 \\
\bottomrule
\end{tabular}
\end{table*}

\begin{table*}
\centering
\caption{Full results of PartedVAE on dsprites for the compositional generalization setting.}
\begin{tabular}{llrrrrrr}
\toprule
       &    &     x &     y &  size &  orient &  shape &   NC \\
\midrule
NC 0-0 & CG linear &  8.78 & 25.76 & 15.76 &    2.47 &   2.52 & 0.00 \\
       & CG & 99.44 & 99.14 &  0.01 &    2.48 &   0.00 & 0.00 \\
NC 5-2 & CG linear & 19.78 & 21.52 &  0.24 &   16.12 &   1.33 & 0.00 \\
       & CG & 16.81 & 20.48 &  0.18 &   36.63 &  60.67 & 0.00 \\
NC 2-1 & CG linear & 50.62 & 49.59 &  0.02 &   26.67 &  98.71 & 0.00 \\
       & CG & 93.12 & 90.44 &  0.00 &   74.15 &  95.71 & 0.00 \\
NC 3-1 & CG linear & 31.28 & 39.57 &  2.21 &    5.88 &  23.20 & 0.00 \\
       & CG & 86.59 & 86.15 &  2.90 &   33.01 &  98.21 & 2.23 \\
NC 3-2 & CG linear & 25.13 & 25.57 &  0.00 &   10.81 &   2.38 & 0.00 \\
       & CG & 27.62 & 22.72 &  0.00 &   13.23 &  60.12 & 0.00 \\
\bottomrule
\end{tabular}
\end{table*}

\begin{table*}
\centering
\caption{Full results of PartedVAE-semisupervised on dsprites for the compositional generalization setting.}
\begin{tabular}{llrrrrrr}
\toprule
       &    &     x &     y &  size &  orient &  shape &   NC \\
\midrule
NC 0-0 & CG linear & 43.96 & 26.48 & 13.41 &    2.95 &   7.58 & 0.00 \\
       & CG & 80.95 & 87.81 &  1.19 &    4.81 &  65.34 & 0.07 \\
NC 5-2 & CG linear & 19.61 & 19.25 &  0.00 &    8.70 &   0.00 & 0.00 \\
       & CG & 15.49 & 20.46 &  0.07 &    8.85 &  14.52 & 0.00 \\
NC 2-1 & CG linear & 53.81 & 69.43 &  1.07 &   15.81 &  39.05 & 0.00 \\
       & CG & 94.84 & 93.81 &  0.23 &   50.94 &  99.83 & 0.23 \\
NC 3-1 & CG linear & 42.65 & 52.21 &  3.89 &   11.40 &  53.01 & 0.00 \\
       & CG & 87.30 & 83.14 &  4.55 &   29.54 &  92.29 & 2.94 \\
NC 3-2 & CG linear & 25.22 & 25.84 &  0.19 &    7.90 &  15.72 & 0.00 \\
       & CG & 21.01 & 25.08 &  0.02 &   14.37 &  53.86 & 0.00 \\
\bottomrule
\end{tabular}
\end{table*}

\begin{table*}
\centering
\caption{Full results of WeakDE on dsprites for the compositional generalization setting.}
\begin{tabular}{llrrrrrr}
\toprule
       &    &     x &     y &  size &  orient &  shape &   NC \\
\midrule
NC 0-0 & CG linear & 67.06 & 55.46 & 28.83 &    2.75 &   0.00 & 0.00 \\
       & CG & 95.00 & 93.37 &  4.57 &   13.08 &  13.14 & 0.54 \\
NC 5-2 & CG linear & 30.97 & 30.00 &  0.08 &    2.95 &   0.00 & 0.00 \\
       & CG & 28.75 & 32.85 &  0.00 &   42.50 &  65.22 & 0.00 \\
NC 2-1 & CG linear & 45.72 & 45.86 & 15.60 &    3.80 &   0.24 & 0.00 \\
       & CG & 90.76 & 91.81 &  3.38 &   50.05 & 100.00 & 3.38 \\
NC 3-1 & CG linear & 32.40 & 39.59 &  5.65 &    3.82 &   0.43 & 0.01 \\
       & CG & 89.90 & 90.93 &  2.35 &   50.28 &  99.92 & 2.33 \\
NC 3-2 & CG linear & 32.24 & 35.32 &  2.93 &    3.03 &  16.48 & 0.00 \\
       & CG & 50.35 & 49.95 &  0.00 &   60.74 &  97.33 & 0.00 \\
\bottomrule
\end{tabular}
\end{table*}

\begin{table*}
\centering
\caption{Full results of $\beta$-VAE on 3dshapes for the compositional generalization setting.}
\begin{tabular}{llrrrrrrr}
\toprule
       &    &  floor h &  wall h &  size &  orient &  obj h &  shape &    NC \\
\midrule
NC 0-0 & CG linear &    99.25 &   99.03 & 48.31 &   86.55 &   0.00 &  96.83 &  0.00 \\
       & CG &   100.00 &  100.00 & 46.61 &  100.00 &  52.34 &  99.89 &  4.97 \\
NC 5-2 & CG linear &   100.00 &  100.00 &  0.00 &   92.12 & 100.00 & 100.00 &  0.00 \\
       & CG &   100.00 &  100.00 &  0.00 &  100.00 & 100.00 & 100.00 &  0.00 \\
NC 2-1 & CG linear &   100.00 &   99.87 &  0.00 &   64.17 &  16.77 & 100.00 &  0.00 \\
       & CG &   100.00 &  100.00 & 38.27 &   99.66 &  79.52 & 100.00 & 32.94 \\
NC 3-1 & CG linear &    99.41 &   99.70 &  0.00 &   90.09 &  41.61 & 100.00 &  0.00 \\
       & CG &   100.00 &  100.00 & 47.07 &  100.00 &  15.80 & 100.00 &  0.00 \\
NC 3-2 & CG linear &    99.95 &   99.93 & 15.86 &   90.67 &   0.00 &  99.33 &  0.00 \\
       & CG &   100.00 &  100.00 &  0.00 &   99.77 & 100.00 & 100.00 &  0.00 \\
NC 7-3 & CG linear &    97.30 &   96.90 & 59.35 &   78.43 &   0.00 & 100.00 &  0.00 \\
       & CG &   100.00 &   99.88 & 43.79 &   83.65 &  93.99 & 100.00 & 37.78 \\
NC 0-3 & CG linear &    99.59 &   99.81 &  0.00 &   89.38 &  63.08 &  68.14 &  0.00 \\
       & CG &   100.00 &  100.00 & 53.88 &  100.00 &  44.76 &  67.69 & 14.41 \\
NC 7-1 & CG linear &    97.33 &   98.96 & 11.47 &   94.43 &   0.00 & 100.00 &  0.00 \\
       & CG &   100.00 &  100.00 & 59.56 &  100.00 &  79.87 & 100.00 & 49.19 \\
\bottomrule
\end{tabular}
\end{table*}

\begin{table*}
\centering
\caption{Full results of $\beta$-TCVAE on 3dshapes for the compositional generalization setting.}
\begin{tabular}{llrrrrrrr}
\toprule
       &    &  floor h &  wall h &  size &  orient &  obj h &  shape &    NC \\
\midrule
NC 0-0 & CG linear &    93.59 &   95.03 & 45.30 &   90.57 &  76.21 &  76.68 & 26.85 \\
       & CG &    99.11 &  100.00 & 45.35 &   99.96 &  49.21 &  80.89 &  0.03 \\
NC 5-2 & CG linear &    99.12 &   98.65 &  1.41 &   97.91 &  95.38 &  90.47 &  1.41 \\
       & CG &   100.00 &  100.00 &  0.00 &  100.00 & 100.00 &  93.93 &  0.00 \\
NC 2-1 & CG linear &   100.00 &  100.00 &  0.00 &   88.60 &  74.31 &  99.99 &  0.00 \\
       & CG &   100.00 &  100.00 & 43.31 &   99.95 &  73.91 & 100.00 & 36.43 \\
NC 3-1 & CG linear &    99.85 &   99.88 &  0.00 &   83.13 &  84.91 & 100.00 &  0.00 \\
       & CG &   100.00 &  100.00 & 12.79 &  100.00 &  84.64 & 100.00 &  0.00 \\
NC 3-2 & CG linear &    99.04 &   99.73 &  0.00 &   85.46 &  99.99 &  99.87 &  0.00 \\
       & CG &   100.00 &  100.00 &  0.00 &   99.69 & 100.00 &  99.93 &  0.00 \\
NC 7-3 & CG linear &    99.99 &   99.21 &  8.26 &   83.51 &  25.83 & 100.00 &  2.92 \\
       & CG &   100.00 &   99.99 & 39.67 &   97.91 &  90.04 & 100.00 & 30.63 \\
NC 0-3 & CG linear &    99.94 &  100.00 &  0.04 &   87.69 &   4.49 &  69.07 &  0.00 \\
       & CG &   100.00 &  100.00 & 22.02 &  100.00 &  53.25 &  81.16 &  2.39 \\
NC 7-1 & CG linear &    99.70 &   99.65 &  1.99 &   70.85 &  30.21 & 100.00 &  0.00 \\
       & CG &   100.00 &   99.92 & 97.39 &   99.99 &  94.09 & 100.00 & 92.89 \\
\bottomrule
\end{tabular}
\end{table*}

\begin{table*}
\centering
\caption{Full results of FactorVAE on 3dshapes for the compositional generalization setting.}
\begin{tabular}{llllrrrrr}
\toprule
       &    & floor h & wall h &  size &  orient &  obj h &  shape &    NC \\
\midrule
NC 0-0 & CG linear &     100 &   99.8 &  0.00 &   65.11 &  47.82 &  48.82 &  0.00 \\
       & CG &     100 &    100 & 38.69 &  100.00 &  40.93 &  99.16 &  0.73 \\
NC 5-2 & CG linear &   93.81 &  95.39 &  0.00 &   59.65 &  56.00 &  77.95 &  0.00 \\
       & CG &     100 &    100 &  0.00 &   99.97 & 100.00 & 100.00 &  0.00 \\
NC 2-1 & CG linear &   88.57 &  90.95 &  0.00 &   33.27 &   2.14 &  61.83 &  0.00 \\
       & CG &       - &      - &  1.92 &   99.32 &  51.35 & 100.00 &  0.91 \\
NC 3-1 & CG linear &   93.55 &  88.59 &  0.00 &   56.30 &   0.00 &  65.76 &  0.00 \\
       & CG &     100 &    100 & 49.81 &  100.00 &  98.73 & 100.00 & 48.74 \\
NC 3-2 & CG linear &     100 &    100 &  0.00 &   83.40 &  99.06 & 100.00 &  0.00 \\
       & CG &     100 &    100 & 11.51 &   99.80 &  99.90 & 100.00 & 11.51 \\
NC 7-3 & CG linear &   91.25 &  89.19 & 13.79 &   51.55 &   0.00 &  97.49 &  0.00 \\
       & CG &     100 &  95.83 & 26.48 &   94.97 &  67.05 & 100.00 &  2.47 \\
NC 0-3 & CG linear &   98.27 &  99.94 & 14.73 &   58.26 &  61.55 &  90.27 &  0.32 \\
       & CG &       0 &    100 & 22.18 &   99.63 &  56.30 &  99.41 &  0.95 \\
NC 7-1 & CG linear &     100 &    100 & 79.76 &   98.46 &  28.47 &  99.78 &  8.97 \\
       & CG &     100 &    100 & 79.76 &   98.46 &  28.47 &  99.78 &  8.97 \\
\bottomrule
\end{tabular}
\end{table*}

\begin{table*}
\centering
\caption{Full results of PartedVAE on 3dshapes for the compositional generalization setting.}
\begin{tabular}{llrrrrrrr}
\toprule
       &    &  floor h &  wall h &  size &  orient &  obj h &  shape &    NC \\
\midrule
NC 0-0 & CG linear &    92.95 &   91.58 &  0.75 &   76.03 &  80.19 &   0.00 &  0.43 \\
       & CG &    99.07 &   99.04 & 31.42 &   99.35 &  92.53 &  67.45 & 27.71 \\
NC 5-2 & CG linear &    98.63 &   91.13 &  0.00 &   89.03 &  94.97 &  13.33 &  0.00 \\
       & CG &     0.00 &   99.34 &  2.49 &   99.71 &  99.87 &  99.93 &  2.49 \\
NC 2-1 & CG linear &    91.69 &   93.41 & 57.20 &   85.97 &  97.07 &   0.00 & 55.72 \\
       & CG &    99.87 &   99.12 & 94.90 &   99.71 &  99.57 &  64.74 & 94.53 \\
NC 3-1 & CG linear &    96.72 &   91.91 &  0.00 &   83.17 &  86.87 &  99.97 &  0.00 \\
       & CG &    99.91 &   97.51 &  0.00 &   99.07 &  99.82 &  99.94 &  0.00 \\
NC 3-2 & CG linear &    93.36 &   98.02 &  0.00 &   86.93 &  89.82 &  92.88 &  0.00 \\
       & CG &    99.55 &   99.75 &  1.97 &   99.49 &  99.69 &  99.95 &  1.97 \\
NC 7-3 & CG linear &    94.49 &   89.35 &  0.00 &   67.20 &  71.25 &   0.64 &  0.00 \\
       & CG &    98.38 &   96.93 & 18.09 &   67.31 &  96.62 &  98.57 & 17.62 \\
NC 0-3 & CG linear &    98.17 &   97.79 &  0.00 &   80.65 &  80.35 &  28.11 &  0.00 \\
       & CG &    99.09 &   98.93 &  0.13 &   97.96 &  73.55 &  24.63 &  0.01 \\
NC 7-1 & CG linear &    94.46 &   98.07 & 21.74 &   86.32 &  98.42 &   0.00 & 21.72 \\
       & CG &    98.29 &   99.45 & 28.81 &   98.67 &  99.91 &  85.17 & 28.81 \\
\bottomrule
\end{tabular}
\end{table*}

\begin{table*}
\centering
\caption{Full results of PartedVAE on 3dshapes for the compositional generalization setting.}
\begin{tabular}{llrrrrrrr}
\toprule
       &    &  floor h &  wall h &  size &  orient &  obj h &  shape &    NC \\
\midrule
NC 0-0 & CG linear &    84.14 &   81.98 &  2.17 &   15.61 &   0.80 &  86.56 &  0.00 \\
       & CG &    98.41 &   98.77 & 23.85 &   20.11 &  71.33 &  97.00 & 13.74 \\
NC 5-2 & CG linear &    97.77 &   97.29 &  0.00 &   88.28 &   5.29 &  91.21 &  0.00 \\
       & CG &    98.98 &   99.91 & 30.91 &   99.23 &  99.21 &  99.17 & 30.91 \\
NC 2-1 & CG linear &    99.96 &   92.79 &  0.00 &   88.49 &  19.61 &  91.82 &  0.00 \\
       & CG &    99.99 &   99.35 & 30.35 &   99.49 & 100.00 &  99.22 & 30.35 \\
NC 3-1 & CG linear &    99.61 &   90.49 &  0.00 &   23.73 &  83.33 &  93.39 &  0.00 \\
       & CG &    99.91 &   99.11 & 38.73 &   73.62 &  98.71 &  99.65 & 38.31 \\
NC 3-2 & CG linear &    93.93 &   98.53 &  6.41 &   38.83 &  99.09 &  94.86 &  5.69 \\
       & CG &    99.87 &   99.36 & 95.13 &   65.39 &  99.93 &  99.45 & 95.06 \\
NC 7-3 & CG linear &    98.71 &   93.34 &  0.00 &   72.27 &  18.47 &  89.05 &  0.00 \\
       & CG &    97.44 &   99.29 & 37.93 &   83.13 &  85.01 &  97.67 & 37.31 \\
NC 0-3 & CG linear &    83.18 &   89.81 &  0.06 &   45.12 &  44.21 &  80.80 &  0.03 \\
       & CG &    83.29 &   99.48 &  4.13 &   57.53 &  23.19 &  81.03 &  1.50 \\
NC 7-1 & CG linear &    70.73 &   83.83 & 45.95 &   13.14 &  21.61 &  84.67 &  2.12 \\
       & CG &    97.48 &   99.06 & 30.89 &   20.65 &  56.57 &  98.29 & 12.87 \\
\bottomrule
\end{tabular}
\end{table*}

\begin{table*}
\centering
\caption{Full results of WeakDE on 3dshapes for the compositional generalization setting.}
\begin{tabular}{llrrrrrrr}
\toprule
       &    &  floor h &  wall h &  size &  orient &  obj h &  shape &    NC \\
\midrule
NC 0-0 & CG linear &    72.72 &   76.91 &  6.35 &   47.21 &  82.32 &  65.43 &  2.61 \\
       & CG &    78.71 &   77.17 & 44.06 &   73.61 &  82.43 &  79.39 & 33.00 \\
NC 5-2 & CG linear &    95.11 &   95.78 &  0.00 &   67.89 &  28.85 &  94.55 &  0.00 \\
       & CG &    94.98 &   94.80 &  2.63 &   90.65 &  99.87 &  95.61 &  2.63 \\
NC 2-1 & CG linear &    97.13 &   96.12 &  0.00 &   61.19 &  37.03 &  93.30 &  0.00 \\
       & CG &    97.71 &   96.29 & 19.42 &   89.35 &  97.95 &  96.95 & 19.32 \\
NC 3-1 & CG linear &    93.30 &   94.49 &  0.07 &   93.72 &  88.01 &  91.71 &  0.01 \\
       & CG &    92.23 &   93.37 &  0.00 &   98.49 &  99.52 &  94.33 &  0.00 \\
NC 3-2 & CG linear &    97.22 &   95.55 &  0.00 &   75.38 &   3.08 &  92.79 &  0.00 \\
       & CG &    95.69 &   94.25 &  8.77 &   92.27 &  99.25 &  96.82 &  8.67 \\
NC 7-3 & CG linear &    88.71 &   95.09 &  0.46 &   12.38 &  48.32 &  92.52 &  0.00 \\
       & CG &    87.91 &   95.92 &  6.57 &   68.71 & 100.00 &  94.47 &  6.57 \\
NC 0-3 & CG linear &    97.44 &   94.11 &  6.71 &   17.36 &  59.99 &  88.23 &  0.40 \\
       & CG &    98.04 &   96.01 & 22.81 &   84.21 &  91.52 &  93.67 & 15.16 \\
NC 7-1 & CG linear &    92.85 &   89.84 &  0.61 &   71.59 &   0.28 &  92.05 &  0.00 \\
       & CG &    89.60 &   87.49 &  0.29 &   78.94 &  97.97 &  93.85 &  0.29 \\
\bottomrule
\end{tabular}
\end{table*}

\begin{table*}
\centering
\caption{Full results of $\beta$-VAE on mpi3d for the compositional generalization setting.}
\begin{tabular}{llrrrrrrrr}
\toprule
       &    &   hor &  vert &  size &  obj h &  shape &  cam he &   bg h &    NC \\
\midrule
NC 0-0 & CG linear &  9.68 &  7.66 & 80.89 &  95.20 &   0.13 &   99.37 &  38.20 &  0.00 \\
       & CG & 75.16 & 55.28 & 99.86 &  98.63 &  20.23 &  100.00 & 100.00 & 20.12 \\
NC 1-4 & CG linear & 16.50 &  6.34 & 42.34 &  91.57 &   5.06 &   99.00 &  99.85 &  0.00 \\
       & CG & 51.51 & 37.19 & 94.21 &  99.93 &  17.16 &  100.00 &   0.00 & 15.56 \\
NC 0-5 & CG linear & 12.55 & 12.39 & 83.41 &  90.54 &  26.06 &   99.98 &  50.46 & 25.62 \\
       & CG & 56.62 & 47.98 & 93.39 &  99.97 &  21.40 &  100.00 & 100.00 & 15.56 \\
NC 1-3 & CG linear & 16.35 & 10.47 & 67.32 &  81.67 &   0.00 &   99.36 &  57.25 &  0.00 \\
       & CG & 87.19 & 65.85 & 99.96 &  99.98 &   0.62 &  100.00 & 100.00 &  0.62 \\
NC 1-1 & CG linear & 15.79 & 10.85 & 92.48 &  74.80 &   0.00 &   97.95 &  55.88 &  0.00 \\
       & CG & 73.84 & 56.43 & 99.98 &  99.77 &   7.93 &  100.00 &  99.70 &  7.91 \\
NC 0-2 & CG linear & 16.22 & 10.56 & 70.52 &  80.97 &   0.00 &   99.94 &  46.48 &  0.00 \\
       & CG & 86.08 & 67.12 & 99.24 &  99.98 &   0.55 &  100.00 & 100.00 &  0.52 \\
\bottomrule
\end{tabular}
\end{table*}

\begin{table*}
\centering
\caption{Full results of $\beta$-TCVAE on mpi3d for the compositional generalization setting.}
\begin{tabular}{llrrrrrrrr}
\toprule
       &    &   hor &  vert &  size &  obj h &  shape &  cam he &   bg h &    NC \\
\midrule
NC 0-0 & CG linear & 15.62 &  9.01 & 97.71 &  40.08 &   0.03 &  100.00 &  53.47 &  0.01 \\
       & CG & 69.87 & 53.41 & 99.87 &  96.95 &   2.30 &  100.00 & 100.00 &  2.30 \\
NC 1-4 & CG linear & 18.32 & 12.63 & 68.95 &  72.15 &   1.31 &   99.99 &  45.06 &  0.00 \\
       & CG & 54.29 & 40.48 & 93.83 &  99.66 &  13.79 &  100.00 & 100.00 & 11.74 \\
NC 0-5 & CG linear &  6.51 & 10.99 & 93.31 &  38.70 &  20.30 &  100.00 &  44.35 & 20.30 \\
       & CG & 51.64 & 41.37 & 88.87 &  95.26 &  10.13 &  100.00 &  99.98 &  5.80 \\
NC 1-3 & CG linear & 15.55 &  8.44 & 86.18 &  60.64 &   0.00 &   99.98 &  45.47 &  0.00 \\
       & CG & 85.38 & 64.75 & 99.97 &  99.69 &   0.53 &  100.00 &  98.46 &  0.53 \\
NC 1-1 & CG linear & 12.76 &  8.45 & 78.44 &  74.00 &   0.00 &   99.18 &  44.43 &  0.00 \\
       & CG & 67.06 & 51.68 & 99.90 &  99.53 &  11.52 &  100.00 & 100.00 & 11.47 \\
NC 0-2 & CG linear & 18.69 &  9.13 & 73.26 &  78.03 &   0.00 &  100.00 &  44.74 &  0.00 \\
       & CG & 89.86 & 70.40 & 98.49 &  99.76 &   0.24 &  100.00 &  99.99 &  0.20 \\
\bottomrule
\end{tabular}
\end{table*}

\begin{table*}
\centering
\caption{Full results of FactorVAE on mpi3d for the compositional generalization setting.}
\begin{tabular}{llrrrrrrrr}
\toprule
       &    &   hor &  vert &  size &  obj h &  shape &  cam he &  bg h &    NC \\
\midrule
NC 0-0 & CG linear & 16.06 &  8.90 & 71.63 &  30.92 &   6.94 &   97.43 & 51.25 &  4.30 \\
       & CG & 65.00 & 50.01 & 99.57 &  93.45 &   9.92 &   99.95 & 99.92 &  9.71 \\
NC 1-4 & CG linear & 14.03 & 11.09 & 40.42 &  57.51 &   0.05 &   99.86 & 42.14 &  0.00 \\
       & CG & 46.42 & 34.33 & 93.18 &  98.03 &  29.25 &  100.00 & 99.99 & 26.63 \\
NC 0-5 & CG linear & 11.64 &  7.73 & 92.40 &  64.68 &   8.43 &   99.85 & 52.75 &  8.43 \\
       & CG & 58.01 & 39.48 & 97.24 &  98.22 &   3.66 &  100.00 & 99.97 &  2.70 \\
NC 1-3 & CG linear & 15.18 &  9.77 & 57.91 &  83.34 &   0.00 &   99.33 & 46.93 &  0.00 \\
       & CG & 79.48 & 60.41 & 99.89 &  99.94 &   0.55 &  100.00 & 99.99 &  0.55 \\
NC 1-1 & CG linear & 17.13 & 11.69 & 74.08 &  76.06 &   0.00 &   98.12 & 46.27 &  0.00 \\
       & CG & 58.10 & 47.78 & 98.25 &  98.80 &  10.18 &  100.00 & 99.32 &  9.44 \\
NC 0-2 & CG linear & 16.27 & 11.76 & 67.23 &  80.28 &   0.25 &   99.08 & 48.57 &  0.00 \\
       & CG & 78.80 & 59.93 & 98.58 &  99.75 &   0.62 &  100.00 & 99.98 &  0.55 \\
\bottomrule
\end{tabular}
\end{table*}

\begin{table*}
\centering
\caption{Full results of PartedVAE on mpi3d for the compositional generalization setting.}
\begin{tabular}{llrrrrrrrr}
\toprule
       &    &   hor &  vert &  size &  obj h &  shape &  cam he &  bg h &   NC \\
\midrule
NC 0-0 & CG linear &  9.29 &  2.77 & 93.64 &  23.70 &    100 &   92.70 &  0.00 & 0.00 \\
       & CG & 21.86 & 18.07 & 98.32 &  70.95 &    100 &   99.57 &  0.06 & 0.06 \\
NC 1-4 & CG linear & 12.39 &  5.12 & 60.38 &  38.19 &    100 &   97.27 &  0.00 & 0.00 \\
       & CG & 27.80 & 18.19 & 80.16 &  85.51 &    100 &   99.52 &  0.08 & 0.07 \\
NC 0-5 & CG linear & 10.58 &  4.20 & 88.16 &  31.08 &    100 &   99.21 &  0.00 & 0.00 \\
       & CG & 28.58 & 24.29 & 92.93 &  90.35 &    100 &   99.99 &  0.05 & 0.04 \\
NC 1-3 & CG linear & 10.65 &  4.67 & 90.47 &  31.14 &    100 &   96.15 &  0.00 & 0.00 \\
       & CG & 28.40 & 17.32 & 97.09 &  86.12 &    100 &   98.58 &  0.95 & 0.85 \\
NC 1-1 & CG linear & 11.27 &  4.79 & 84.14 &  31.40 &    100 &   98.36 &  0.00 & 0.00 \\
       & CG & 31.14 & 18.25 & 92.75 &  88.23 &    100 &   99.66 &  0.02 & 0.01 \\
NC 0-2 & CG linear & 12.43 &  4.90 & 59.53 &  37.04 &    100 &   97.06 &  0.00 & 0.00 \\
       & CG & 29.03 & 18.11 & 79.58 &  86.12 &    100 &   99.48 &  0.05 & 0.01 \\
\bottomrule
\end{tabular}
\end{table*}

\begin{table*}
\centering
\caption{Full results of PartedVAE-semisupervised on mpi3d for the compositional generalization setting.}
\begin{tabular}{llrrrrrrrr}
\toprule
       &    &   hor &  vert &  size &  obj h &  shape &  cam he &  bg h &   NC \\
\midrule
NC 0-0 & CG linear & 11.81 &  4.95 & 78.04 &  25.77 &   0.00 &  100.00 & 42.80 & 0.00 \\
       & CG & 23.69 & 20.91 & 97.15 &  76.42 &   0.28 &  100.00 & 51.88 & 0.17 \\
NC 1-4 & CG linear & 12.78 &  5.83 & 50.73 &  44.75 &   0.00 &  100.00 & 46.50 & 0.00 \\
       & CG & 16.32 & 12.89 & 54.24 &  90.76 &   0.50 &  100.00 & 64.07 & 0.02 \\
NC 0-5 & CG linear & 11.32 &  4.36 & 80.33 &  39.31 &   0.00 &  100.00 & 86.30 & 0.00 \\
       & CG & 11.32 &  4.36 & 80.33 &  39.31 &   0.00 &  100.00 & 86.30 & 0.00 \\
NC 1-3 & CG linear & 13.23 &  7.50 & 84.65 &  37.04 &   0.00 &   99.98 & 76.83 & 0.00 \\
       & CG & 51.80 & 29.91 & 98.30 &  93.74 &   0.09 &  100.00 & 98.23 & 0.07 \\
NC 1-1 & CG linear & 10.27 &  5.09 & 74.54 &  30.44 &   0.00 &   96.35 & 56.67 & 0.00 \\
       & CG & 29.27 & 18.07 & 96.88 &  92.19 &   0.49 &   99.96 & 94.38 & 0.31 \\
NC 0-2 & CG linear & 11.04 &  4.42 & 52.11 &  47.71 &   0.00 &  100.00 & 34.08 & 0.00 \\
       & CG & 28.44 & 20.42 & 73.13 &  88.58 &   0.04 &  100.00 & 38.63 & 0.01 \\
\bottomrule
\end{tabular}
\end{table*}

\begin{table*}
\centering
\caption{Full results of WeakDE on mpi3d for the compositional generalization setting.}
\begin{tabular}{llrrrrrrrr}
\toprule
       &    &   hor &  vert &  size &  obj h &  shape &  cam he &   bg h &    NC \\
\midrule
NC 0-0 & CG linear & 14.27 & 13.70 & 30.99 &  29.20 &   0.02 &  100.00 &  95.09 &  0.00 \\
       & CG & 63.79 & 51.82 & 98.45 &  91.12 &  14.62 &  100.00 & 100.00 & 13.84 \\
NC 1-4 & CG linear & 16.80 & 12.68 & 32.39 &  52.54 &   0.00 &  100.00 &  64.50 &  0.00 \\
       & CG & 30.78 & 18.92 & 87.65 &  97.88 &   3.69 &  100.00 & 100.00 &  3.13 \\
NC 0-5 & CG linear & 11.16 & 12.20 & 59.33 &  35.83 &   0.00 &  100.00 &  79.64 &  0.00 \\
       & CG & 45.44 & 35.85 & 96.86 &  95.43 &  13.44 &  100.00 & 100.00 & 12.95 \\
NC 1-3 & CG linear & 18.41 & 11.59 & 59.82 &  37.31 &   0.00 &  100.00 &  62.56 &  0.00 \\
       & CG & 75.88 & 54.61 & 99.15 &  98.37 &   0.00 &  100.00 &  99.35 &  0.00 \\
NC 1-1 & CG linear & 17.74 & 21.09 & 74.95 &  44.00 &   0.00 &   99.99 &  90.88 &  0.00 \\
       & CG & 70.73 & 61.00 & 99.89 &  97.19 &  11.80 &  100.00 &  99.89 & 11.77 \\
NC 0-2 & CG linear & 20.49 & 19.73 & 47.85 &  61.64 &   0.20 &  100.00 &  91.91 &  0.00 \\
       & CG & 80.54 & 69.87 & 95.28 &  98.97 &   1.17 &  100.00 & 100.00 &  0.92 \\
\bottomrule
\end{tabular}
\end{table*}

\begin{table*}
\centering
\caption{Full results of MTD on dsprites for the compositional generalization setting.}
\begin{tabular}{lrrrrrr}
\toprule
{} &     x &     y &   size &  orient &  shape &    zs \\
\midrule
NC 5-2            & 77.90 & 77.60 &   6.10 &    3.30 &  33.90 &  2.53 \\
NC 2-1            & 94.10 & 95.80 &  17.80 &    3.10 &  25.60 &  3.79 \\
NC 3-1            & 93.90 & 95.80 &  17.80 &    3.60 &  35.60 &  4.48 \\
NC 3-2            & 88.20 & 88.80 &   3.60 &    3.10 &  38.30 &  0.89 \\
Normal test set 1 & 99.00 & 99.40 & 100.00 &   83.30 & 100.00 & 99.97 \\
Normal test set 2 & 99.00 & 98.60 &  99.80 &   75.00 & 100.00 & 99.80 \\
Normal test set 3 & 98.30 & 98.60 &  99.80 &   75.40 & 100.00 & 99.80 \\
Normal test set 4 & 97.20 & 98.80 &  99.40 &   74.10 & 100.00 & 99.37 \\
Normal test set 5 & 98.70 & 97.40 &  99.60 &   73.00 & 100.00 & 99.58 \\
\bottomrule
\end{tabular}
\end{table*}

\begin{table*}
\centering
\caption{Full results of MTD on 3dshapes for the compositional generalization setting.}
\begin{tabular}{lrrrrrrr}
\toprule
{} &  floor h &  wall h &   size &  orient &  obj h &  shape &     zs \\
\midrule
NC 0-0            &    11.10 &   90.90 &  18.80 &  100.00 &  29.80 &  93.00 &   1.29 \\
NC 5-2            &     8.50 &   92.40 &   4.20 &  100.00 &  26.90 &  94.80 &   0.57 \\
NC 2-1            &     8.20 &   90.50 &  12.20 &  100.00 &  27.80 &  93.90 &   4.25 \\
NC 3-1            &    13.90 &   90.20 &   9.40 &  100.00 &  24.70 &  91.30 &   3.22 \\
NC 3-2            &    10.10 &   96.30 &  10.20 &  100.00 &  24.10 &  94.70 &   2.17 \\
NC 7-3            &    10.09 &   96.02 &  12.79 &   99.94 &  25.03 &  96.35 &   2.00 \\
NC 0-3            &     9.28 &   88.57 &   7.69 &  100.00 &  21.31 &  89.02 &   1.05 \\
NC 7-1            &     9.57 &   96.99 &  15.73 &  100.00 &  23.53 &  96.87 &   0.75 \\
Normal test set 1 &   100.00 &  100.00 & 100.00 &  100.00 & 100.00 & 100.00 & 100.00 \\
Normal test set 2 &   100.00 &  100.00 & 100.00 &  100.00 & 100.00 & 100.00 & 100.00 \\
Normal test set 3 &     9.20 &  100.00 & 100.00 &  100.00 & 100.00 & 100.00 & 100.00 \\
Normal test set 4 &     7.30 &  100.00 & 100.00 &  100.00 & 100.00 & 100.00 & 100.00 \\
Normal test set 5 &    20.00 &  100.00 & 100.00 &  100.00 & 100.00 & 100.00 & 100.00 \\
\bottomrule
\end{tabular}
\end{table*}

\begin{table*}
\centering
\caption{Full results of MTD on mpi3d for the compositional generalization setting.}
\begin{tabular}{lrrrrrrrr}
\toprule
{} &  hor &  vert &  size &  obj h &  shape &  cam he &   bg h &    zs \\
\midrule
NC 0-0            & 2.20 &  2.30 & 50.50 &  21.70 &   8.50 &   43.90 &  58.60 &  2.67 \\
NC 1-4            & 2.50 &  0.00 & 96.60 &  53.90 &  11.60 &   99.60 & 100.00 & 11.40 \\
NC 0-5            & 1.79 &  2.44 & 49.55 &  20.93 &   7.95 &   42.79 &  58.67 &  7.36 \\
NC 1-3            & 2.32 &  2.52 & 51.52 &  19.62 &   9.77 &   42.05 &  57.68 &  3.06 \\
NC 1-1            & 2.92 &  2.56 & 42.73 &  19.97 &   8.56 &   43.30 &  58.35 &  2.90 \\
NC 0-2            & 2.39 &  2.52 & 59.98 &  21.82 &   6.46 &   45.27 &  58.73 &  2.68 \\
Normal test set 1 & 2.50 &  2.50 & 99.90 & 100.00 &  96.20 &  100.00 & 100.00 & 96.10 \\
Normal test set 2 & 3.43 &  2.75 & 99.92 &  99.99 &  95.94 &  100.00 & 100.00 & 95.88 \\
Normal test set 3 & 2.48 &  3.00 & 99.92 &  99.99 &  96.43 &  100.00 & 100.00 & 96.37 \\
Normal test set 4 & 2.70 &  2.34 & 99.94 &  99.99 &  95.56 &  100.00 & 100.00 & 95.51 \\
Normal test set 5 & 2.65 &  2.46 & 99.75 &  99.99 &  95.09 &  100.00 & 100.00 & 94.87 \\
\bottomrule
\end{tabular}
\end{table*}

\end{appendix}
\end{document}